\documentclass[pdflatex,sn-mathphys-num]{sn-jnl}

\usepackage{graphicx}%
\usepackage{multirow, booktabs}
\usepackage{tabularx}
\usepackage{amsmath,amssymb,amsfonts}%
\usepackage{amsthm}%
\usepackage{mathrsfs}%
\usepackage[title]{appendix}%
\usepackage{xcolor}%
\usepackage{textcomp}%
\usepackage{manyfoot}%
\usepackage{booktabs}%
\usepackage{algorithm}%
\usepackage{algorithmicx}%
\usepackage{algpseudocode}%
\usepackage{listings}%
\usepackage{array}
\usepackage{enumitem}
\usepackage{relsize}
\usepackage{bm}
\usepackage{subcaption}
\usepackage{adjustbox}
\usepackage{pgfplots}
\usepackage{tikzscale,tikz-qtree,tikz-qtree-compat}
\usetikzlibrary{matrix,chains,positioning,decorations.pathreplacing,arrows,calc,fit,arrows.meta, shapes.geometric,shapes.misc,shadows}
\usepackage[most]{tcolorbox}

\theoremstyle{thmstyleone}%
\newtheorem{theorem}{Theorem}
\theoremstyle{thmstyletwo}%

\theoremstyle{thmstylethree}%
\newtheorem{definition}{Definition}%
\newtheorem{hypothesis}{Hypothesis}

\renewcommand{\descriptionlabel}[1]{\hspace{\labelsep}\textrm{#1}}

\begin{document}

\title[Framework for the Nature-Inspired Metaheuristics]{Framework for identifying the equivalence between Nature-Inspired Metaheuristics}

\author[1]{\fnm{Iztok} \sur{Fister}}\email{iztok.fister@um.si}

\author[1]{\fnm{Žan} \sur{Hozjan}}\email{zan.hozjan1@student.um.si}

\author*[1]{\fnm{Iztok} \sur{Fister, Jr}}\email{iztok.fister1@um.si}

\author[1]{\fnm{Damjan} \sur{Strnad}}\email{damjan.strnad@um.si}

\affil*[1]{\orgdiv{Faculty of Electrical Engineering and Computer Science}, \orgname{University of Maribor}, \orgaddress{\street{Koro\v{s}ka cesta 46}, \city{Maribor}, \postcode{2000}, \country{Slovenia}}}


\abstract{The domain of metaheuristic optimization has become vibrant due to a flood of new algorithms using a new nature-inspired metaphor but lacking clear methodological novelty. The Criticism behind the development of these algorithms has reached such an extent that the critics started to assert that all novel algorithms are only copies of already developed ones. In this study, we try to show that the situation is not so black and white. Therefore, we define a strong equivalence theorem for estimating the similarity between two nature-inspired metaheuristics, according to which two algorithms are equivalent if, and only if, the cosine similarity of their phenotypic and genotypic feature vectors, characterizing their behavior by searching for the optimal solutions, is above some threshold. On the theorem basis, a framework is developed for identifying the equivalence between nature-inspired metaheuristics. Extensive experimental work using the framework has shown that searching for conditions to achieve the high similarity of the more well-known nature-inspired metaheuristics is hard, or even not possible to achieve, in the limited computational environments.}


\maketitle

\section{Introduction}\label{sec1}

In the domain of metaheuristic optimization, we have been confronted with a flood of metaphor-based algorithms in recent years. Mostly, these algorithms obscure the lack of methodological novelty by using a new Nature-Inspired (NI) metaphor. In fact, the newly proposed algorithms often borrow already known components from the existing nature-inspired metaheuristics~\cite{tzanetos}. This trend was exposed even more after the criticism of S\"{o}rensen~\cite{Sorensen2015Metaheuristics} that caused a lot of discussions confirming his hypothesis~\cite{Lones2014metaheuristic,Campelo2025bestiary}. 

According to the proposal of the new metaphor-based NI algorithms, the number of research papers that critically analyze the mechanisms of newly proposed methods has also increased. Some of the first such papers were written by Weyland in 2010~\cite{weyland2010rigorous} and 2015~\cite{weyland2015critical}, respectively, where the author proved formally that the Harmony Search (HS) algorithm can be considered a special case of evolution strategies. Additionally, Piotrowski et al. in their paper~\cite{piotrowski2014novel}, showed that Black Hole Optimization is, in fact, a simplified version of particle swarm optimization with inertia weight. Other papers that examined these algorithms critically also included studies on Cuckoo Search by Camacho-Villal{\'o}n et al.~\cite{camacho2022analysis}, as well as the Firefly algorithm, Grey Wolf Optimizer, and Bat algorithm~\cite{Camacho2023exposing}. Some recent papers included a critical review of Salp Swarm Optimization (SSO)~\cite{castelli2022salp}. The Grasshopper Optimization Algorithm (GOA) was under scrutiny, while their critics showed that it is not a new algorithm, but can be viewed as a derivative of the particle swarm optimization~\cite{harandi2024grasshopper}.

Unfortunately, the majority of these evidence papers base on theoretical foundations, while how the NI algorithms explore the search space and what results they achieve have not be indicated explicitly. As holds in this domain, a new metaheuristic algorithm is defined already by using a different parameter setting~\cite{eiben2015introduction,Fister2021on}. In line with this, asserting statements like ''The algorithm A is B.'' usually reported in these studies without suitable evidence (e.g., performing extensive ablation studies, exposing circumstances under which the assertion holds, etc.), that justifies this fact generally, seems a little uncritical. This does not mean that the authors of this study do not agree with these theoretical foundations, or even defend the flood of developing the ''novel'' metaphor based NI algorithms. On the contrary, this study is aimed at trying to create the conditions, under which the equivalence of these algorithms could be estimated as fairly as possible from theoretical and practical aspects.

Indeed, the purpose of the study is the expansion and elaboration of the already published paper by \citet{fister2021gecco}, which is threefold:
\begin{itemize}
    \item to set an equivalence theorem for comparing the similarity between two metaheuristic algorithms,
    \item to define the descriptive and similarity metrics for identifying the equivalence of two nature-inspired metaheuristics in the genotype, as well as the phenotype space,
    \item to develop a publicly available framework for identifying the similarity of the NI metaheuristic algorithms based on the equivalence theorem, and measured by statistical similarity metrics and Machine Learning (ML) classification methods.
\end{itemize}
Two types of metrics are defined for identifying the similarity of two NI metaheuristic algorithms: (1) descriptive, and (2) similarity. The former describe the behavior of a particular NI algorithm during one evolutionary run in the phenotype and genotype space, and are collected into the so-called feature vectors. The latter are aimed at identifying the similarity, and are based on a comparison of the feature vectors produced by NI metaheuristics. 

Asserting that two nature-inspired algorithms are equivalent (i.e., expose a similar behavior) demands the fulfilling of two requirements:
\begin{itemize}
    \item obtaining the results of the comparable quality in the phenotype (i.e., solution) space,
    \item describing the equivalent trajectories of the particular individuals by moving through the genotype space affected by the variation operators during all generations.
\end{itemize}
Based on these two requirements, the equivalence theorem is defined in our study as follows:
\begin{theorem}
Two nature-inspired metaheuristic algorithms are equivalent if the results produced in the phenotype space, and the trajectories, described by moving the individuals throughout the genotype space due to acting variation operators in all generations, differ by less than 1~\% regarding the cosine similarity between their corresponding feature vectors.
\label{theorem:1}
\end{theorem}

\noindent In the sense of the equivalence theorem, the following definition needs to be held:

\begin{definition} Two algorithms are \textbf{strongly equivalent} if each element of the feature vector obtained by the first algorithm is matched exactly by the corresponding element of the feature vector produced by the second algorithm, and vice versa.
\label{def:1}
\end{definition}

The proposed framework for identifying the similarity between NI metaheuristic algorithms consists of three components: (1) the control algorithm, (2) a controlled algorithm, and (3) the metaheuristic control system. The control algorithm is used for generating the reference feature vectors, which we then try to reproduce with the controlled algorithm. The controlled algorithm starts with the same initial population (achieved by using the same seed of the random generator). In order to automate the tuning process, a metaheuristic control system similar to \texttt{Meta-GA} as proposed by \citet{Grefenstette1986optimization} is used by the framework. The task of the metaheuristic control system is to search for those hyper-parameter settings of the controlled algorithm, with which the highest cosine similarity is achieved between the feature vectors of the control and controlled algorithm.

Working with the proposed framework demands the following three steps:
\begin{itemize}
\item preprocessing: selecting a set of NI metaheuristics for identifying the strong equivalence between them,
and the algorithm's analysis (i.e., component analysis, parameter analysis, and variation operator design),
\item processing: framework application, 
\item postprocessing: the statistical analysis of the results, and, optionally, visualization of these.
\end{itemize}
The first step refers to selecting the NI algorithms included for analysis with the framework and recognizing their implementation characteristics. While the preprocessing step is more or less theoretically oriented and is aimed for performing by the framework user manually, the remaining two steps are solved automatically using the proposed framework. 

In the mentioned study, the proposed framework was tested on the following set of NI metaheuristic algorithms: (1) accelerated Particle Swarm Optimization (\texttt{accPSO})~\cite{yang2008nature}, (2) accelerated Firefly Algorithm (\texttt{accFA})~\cite{yang2008nature}, (3) original Particle Swarm Optimization (\texttt{PSO})~\cite{kennedy1995particle}, (4) original Firefly Algorithm (\texttt{FA})~\cite{yang2008nature}, and (5) original Bat Algorithm (\texttt{BA})~\cite{yang2010new}. 

Let us assume that someone would like to find the evidence supporting the following four hypotheses: 
\begin{hypothesis}
The \texttt{accPSO} is a strong equivalent to the \texttt{accFA} according to Definition~\ref{def:1}.
\label{hypo:1}
\end{hypothesis}
\begin{hypothesis}
    The metaphors provided the inspiration for the implementation of several metaheuristics components.
\label{hypo:2}
\end{hypothesis}
\begin{hypothesis}
    The conditions are hard to create for achieving the strong equivalence between two NI metaheuristics.
\label{hypo:3}
\end{hypothesis}
\begin{hypothesis}
    The BA is PSO.
\label{hypo:4}
\end{hypothesis}
The purpose of the extensive experimental study was to show whether the support for the above hypotheses could be found by applying the proposed framework. In doing so, the results provided by optimizing the Sphere benchmark function showed that finding the evidence for the posted hypotheses is not easy, in particular with limited computer resources. 

The structure of the remainder of the paper is as follows: Section~~\ref{sec:2} describes the basics of the NI metaheuristic algorithms used in the study. In Section~\ref{sec:3}, the proposed framework for discovering the similarity between NI metaheuristics is discussed according to the equivalence theorem. The performed experiments and the obtained results are the subjects of Section~\ref{sec:4}, while the paper concludes with Section~\ref{sec:5}, where the study findings are summarized and the potential directions are outlined for the future work.    

\section{Metaheuristic Algorithms}\label{sec:2}

\subsection{\texttt{PSO}}

\subsubsection{Nature inspiration for the \texttt{PSO} algorithm}

\texttt{PSO} is a powerful metaheuristic optimization algorithm that is inspired by the swarm behavior observed in nature, such as in fish shoals or bird flocks. The bird flock inspiration lies in the following three rules (the same also applies to the fish shoal):
\begin{enumerate}
    \item The particles in \texttt{PSO} swarm mimic the social behavior of birds by searching for food.
    \item Each bird in a bird flock is determined with a position and particular velocity, while the associated fitness is proportional to the potential amount of feed found at the position.
    \item The birds' directions are adjusted toward spots where they or their neighbors have found more food.
\end{enumerate}
In nature, any of the bird’s observable vicinity is limited to some range. However, having more than one bird allows all the birds to be aware of the larger surface of a fitness function by using the communication within the swarm.  

\subsubsection{Mathematical model of the \texttt{PSO} algorithm}
The \texttt{PSO} translates this natural behavior into a mathematical model. Here, each “bird” is a particle, representing a possible solution to the optimization problem. The position of a particle encodes the solution itself, while its velocity determines how it moves through the search space. Each particle remembers the best position it has found so far (its personal best), and is also influenced by the best position found by any particle in the swarm (the global best).

This dual influence (i.e., self-experience and social learning) creates a dynamic balance between exploration (searching new areas) and exploitation (refining known good areas)~\cite{crepinsek2013exploration}. Just as birds zero in on the best feeding ground collectively, the swarm of particles gradually converges toward an optimal or near-optimal solution.
    
Let $\textbf{x}^{(t)}_i$ and $\textbf{v}^{(t)}_i$ be the position and velocity vectors for the $i$-th particle, respectively. Then, the new velocity vector is expressed by the \texttt{PSO} algorithm as:
\begin{equation}
    v^{(t)}_{i,j}=w\cdot v^{(t)}_{i,j}+c_1\cdot\epsilon_{1,j}(g^*_j-x^{(t)_{i,j}})+c_2\cdot\epsilon_{2,j}((x^*_{i,j}-x^{(t)_{i,j}})),
\end{equation}
where $\mathbf{g}^*$ denotes the global best position, $\mathbf{x}^*_i$ is the local best position, $\boldsymbol\epsilon_1$ and $\boldsymbol\epsilon_2$ are two random vectors whose values are drawn randomly from the uniform distribution in the interval $[0,1]$, the parameter $w$ is the inertial coefficient, and the parameters $c_1$ (also social coefficient) and $c_2$ (also cognitive coefficient) are the learning parameters (also acceleration coefficients), which are, typically, drawn in the interval $[0,2]$. The new position is calculated according to the following equation:
\begin{equation}
    \mathbf{x}^{(t+1)}_i=\mathbf{x}^{(t)}_i+\mathbf{v}^{(t+1)}_i.
    \label{eq:pso}
\end{equation}

The pseudo-code of the \texttt{PSO} is illustrated in Algorithm~\ref{alg:pso}.

\begin{algorithm}[htb]
\caption{The original \texttt{PSO} algorithm}
\label{alg:pso}
\begin{algorithmic}[1]
\Procedure{ParticleSwarmOptimization}{}
\Require inertial coefficient $w$, cognitive $c_1$, social coefficient $c_2$
\State $t\gets 0$;
\State $P^{(0)}\gets$\textsc{Initialize}; \Comment {initialization of population}
\While {$\mathbf{not}$ \textsc{TerminationConditionMeet}}
\ForAll {$\textbf{x}^{(t)}_i\in P^{(t)}$}
\State $f^{(t)}_{i}$ = \textsc{Evaluate}($\mathbf{x}^{(t)}_i$); \Comment {evaluation of candidate solution}
\If {$f^{(t)}_{i} \leq f^{(t)}_{\mathit{best,i}}$}
\State $\mathbf{p}^{(t)}_{i}=\mathbf{x}^{(t)}_{i}$; $f^{(t)}_{\mathit{best,i}}=f^{(t)}_{i}$;
\If {$f^{(t)}_{i} \leq f^{(t)}_{\mathit{g}}$}
\State $\mathbf{g}^{(t)}=\mathbf{x}^{(t)}_{i}$; $f^{(t)}_{\mathit{g}}=f^{(t)}_{i}$;
\EndIf \Comment {preserve the global best solution }
\EndIf \Comment {preserve the local best solution} 
\State $\mathbf{x}^{(t)}_i$ = \textsc{Move}($\mathbf{x}^{(t)}_i,w,c_1,c_2$); \Comment {move the candidate w.r.t. Eq.~(\ref{eq:pso})}
\EndFor
\State $t=t+1$;
\EndWhile 
\EndProcedure
\end{algorithmic}
\end{algorithm}

\subsubsection{Accelerated Particle Swarm Optimization}
In accelerated \texttt{PSO}, the velocity vector is updated using the following simpler equation:
\begin{equation}
    \textbf{v}^{(t+1)}_i=\textbf{v}^{(t)}_i+\alpha\left(\boldsymbol\epsilon-\frac{1}{2}\right)+\beta\left(\textbf{g}^*-\textbf{x}^{(t)}_i\right),
\end{equation}
where $\boldsymbol\epsilon$ is a random variable with values drawn from uniform distribution in the interval $[0,1]$, and $\textbf{g}^*$ is the global best vector. 
Then, the position update is expressed simply as:
\begin{equation}
    \textbf{x}^{(t+1)}_i=\textbf{x}^{(t)}_i+\textbf{v}^{(t+1)}_i.
\end{equation}
In order to improve the convergence, the following equation can be used for updating the position:
\begin{equation}
    \textbf{x}^{(t+1)}_i=\left(1-\beta\right)\textbf{x}^{(t)}_i+\beta\textbf{g}^{*}+\alpha\left(\boldsymbol\epsilon-0.5\right),
    \label{eq:apso}
\end{equation}
where the initial values for $\alpha_0$ are drawn from the interval $\alpha_0\in[0.1,0.4]$, and $\beta_0\in[0.1,0.7]$. 

The final improvement can be made for decreasing the inertial coefficient monotonically using:
\begin{equation}
    \alpha=\alpha_0\exp^{-\gamma t},\qquad\text{or}\qquad\alpha=\alpha_0\gamma^t,
\end{equation}
where $\gamma\in(0,1)$, and $\alpha_0\in [0.5-1]$, e.g., $\alpha=0.7^t$, where $t\in[0,10]$.

\subsection{\texttt{FA}}

\subsubsection{Nature inspiration for the \texttt{FA}}

The \texttt{FA} bases on the following three principles of firefly behavior~\cite{yang2008nature}:
\begin{enumerate}
    \item All fireflies are unisex, so that each firefly is attracted to all the other fireflies regardless of their sex.
    \item The attractiveness of fireflies is proportional to their brightness, meaning that, for any two flashing fireflies, the less bright one is moved toward the brighter one.
    \item The brightness of the firefly is affected by the landscape of the objective function.
\end{enumerate}
In nature, only adult female fireflies are capable of a light emission, due to the chemical process of bioluminescence. The light emission is employed by female fireflies either to attract their potential mates, or to trap their males as prey. As is evident from these facts, the main principles of the FA, as proposed by Yang in~\cite{yang2008nature}, are coupled very loosely to the nature metaphor itself.

\subsubsection{Mathematical model of the FA}
The mathematical model of the FA is defined as: Let $\textbf{x}^{(t)}_i$ be the position vector for the $i$-th firefly. The movement of the $i$-th firefly is affected by a more attractive $j$-th firefly according to the following equation:
\begin{equation}
    \mathbf{x}^{(t+1)}_i=\mathbf{x}^{(t)}_i+\underbrace{\beta_0\exp^{-\gamma r^2_{i,j}}}_{=\beta(r)}(\mathbf{x}^{(t)}_j-\mathbf{x}^{(t)}_i)+\alpha\left(\text{rand}-\frac{1}{2}\right),
    \label{eq:fa}
\end{equation}
where $\beta_0$ denotes the initial attractiveness, $\alpha$ is a randomization factor, $\gamma$ is the absorption coefficient, and $r_{i,j}$ is the 2-D Euclidean distance between two fireflies 
and, theoretically, $\gamma\in[9,\infty)$.  The hyperparameter $\gamma$ is a positive value, which is in practice usually selected from the range $[0.1,10]$. Typically, the randomization factor $alpha$ is drawn from a uniform distribution in the interval $[0,1]$, and the attractiveness function is expressed as $\beta(r)=\beta_0\exp^{-\gamma r^2_{i,j}}$. 

The pseudo-code of the \texttt{FA} is illustrated in Algorithm~\ref{alg:fa}. 

\begin{algorithm}[htb]
\caption{Original Firefly algorithm}
\label{alg:fa}
\begin{algorithmic}[1]
\Procedure{\textsc{FireflyAlgorithm}}{}
\Require randomization $\alpha$, attractiveness $\beta_0$, absorption coefficient $\gamma$
\State $t = 0; \mathbf{x}_{\mathit{best}} = \emptyset;$ \Comment {initialization of parameters}
\State $P^{(0)}$ = \textsc{Initialize}; \Comment {initialization of population}
\While {$\mathbf{not}$ \textsc{TerminationConditionMeet}}
\State \textsc{Evaluate}($P^{(t)}, f(\mathbf{x})$); \Comment {evaluate $\mathbf{x}_i$ regarding the $f(\mathbf{x}_i)$}
\State $\mathbf{x}_{\mathit{best}}$=\textsc{FindTheBestSolution}($P^{(t)})$; \Comment {determine the best solution} 
\ForAll{$\mathbf{x}_i \in P^{(t)}$}
\ForAll{$\mathbf{x}_j \in P^{(t)}$}
\If {$f(\mathbf{x}_j) > f(\mathbf{x}_i)$}
\State $r$=\textsc{Euclidean}$(\mathbf{x}_i,\mathbf{x}_j)$; \Comment calculate Euclidean distance 
\State $\beta = \beta_0*\exp^{-\gamma r^2}$; \Comment{update attractiveness}
\State $\mathbf{x}_i$=\textsc{Move}$(\mathbf{x}_i,\mathbf{x}_j,\alpha,\beta)$; \Comment {move firefly $\mathbf{x}_i$ according Eq.~(\ref{eq:fa})}
\EndIf
\EndFor
\EndFor
\State $\alpha = \alpha*0.98$; \Comment{decay factor - tweak as needed}
\State $t = t+1$;
\EndWhile
\EndProcedure
\end{algorithmic}
\end{algorithm}

\subsubsection{Accelerated Firefly Algorithm}
There are two limiting cases if the original \texttt{FA} is taken into consideration, i.e., (1) when $\gamma\rightarrow 0$, and (2) when $\gamma\rightarrow \infty$. In the former case, the attractiveness $\beta=\beta_0\exp^{-\gamma r^2_{i,j}}$ is constant $\beta=\beta_0$ and $\Gamma=\gamma^{-1/m}\rightarrow \infty$. This corresponds to a definition of the \texttt{accPSO}, because, in that case, Eq.~(\ref{eq:fa}) transforms to the \texttt{accFA}, as follows: 
\begin{equation}
    \mathbf{x}^{(t+1)}_i=\mathbf{x}^{(t)}_i+\beta\left(\mathbf{x}^{*}-\mathbf{x}^{(t)}_i\right)+\alpha\left(\text{rand}-\frac{1}{2}\right),
    \label{eq:afa}
\end{equation}
where vector $\mathbf{x}^{*}$ denotes the global best solution. Let us mention that Eq.~(\ref{eq:afa}) is directly translatable to Eq.~(\ref{eq:apso}) by a simple mathematical rearrangement of the terms. In the latter case,
the attractiveness leads to $\beta(r)\rightarrow \delta(r)$ (i.e., the Dirac delta function) and characteristic distance $\Gamma\rightarrow 0$. This corresponds to the completely random search method.

Theoretically, we could speculate that the \texttt{accPSO} and \texttt{accFA} are strongly equivalent algorithms although they are founded on different NI metaphors.

\subsection{\texttt{BA}}

\subsubsection{Nature inspiration for the \texttt{BA}}
The \texttt{BA} is a metaheuristic optimization algorithm inspired by the echolocation behavior of bats. The algorithm was first introduced by \citet{yang2010new} in 2010, and is based on the following idealized rules:
\begin{enumerate}
\item Bats use echolocation to detect prey and navigate through their environment. 
\item Bats fly randomly with a velocity $\mathbf{v}_i$ at position $\mathbf{x}_i$ using the ultrasound of frequency $Q_{i}$, and loudness $A_0$ to search for prey.
\item They can adjust the frequency (or wavelength) of their emitted pulses and the rate of pulse emission $r$ in the range $[0,1]$, depending on the proximity of their target.
\end{enumerate}
Interestingly, this metaphor also stays on shaky foundations, because the use of echolocation for navigation and finding prey is a characteristic mainly for microbats.

\subsubsection{Mathematical model of the \texttt{BA}}

The mathematical formulation of the \texttt{BA} uses the following equation to update the position $\mathbf{x}_i$ and velocity $\mathbf{v}_i$ of the $i$-th bat at time step $t$:
\begin{equation}
    \mathbf{v}^{(t+1)}_i=\mathbf{v}^{(t)}_i+(\mathbf{x}^{(t)}_i-\mathbf{x}^{(t)}_{best})\cdot Q^{(t)}_i,
    \label{eq:bat}
\end{equation}
and
\begin{equation}
    \mathbf{x}^{(t+1)}_i=\mathbf{x}^{(t)}_i+\mathbf{v}^{(t+1)}_i,
\end{equation}
where $\mathbf{x}^{(t)}_{best}$ is the current global best solution, and $Q_i$ is the frequency of the $i$-th bat, which is calculated as:
\begin{equation}
    Q^{(t)}_i=Q_{\min}+(Q_{\max}-Q_{\min})\cdot \beta^{(t)}_i. 
\end{equation}
Here, $\beta^{(t)}_i$ is a random number drawn from a uniform distribution on the interval $[0,1]$. The loudness $A^{(t)}_i$ and pulse emission rate $r^{(t)}_i$ are updated according to the following equations:
\begin{equation}
    \begin{aligned}
        A^{(t+1)}_i&= \alpha\cdot A^{(t)}_i,\\
        r^{(t+1)}_i&= r^{(t)}_i\cdot(1-\exp^{-\gamma t}), 
    \end{aligned}
\end{equation}
where the hyper-parameters $\alpha\in [0.97,0.99]$ and $\gamma\in[0,10]$ are constants. 

Two hybrid components are embedded into the \texttt{BA} algorithm as follows: (1) Simulated Annealing (\texttt{SA}) is a probabilistic technique for approximating the global optimum of a given function~~\cite{Kirkpatrick1983optimization}, and (2) Gaussian Random Walk (\texttt{GRW}) is a stochastic process describing a path that consists of a succession of random steps in some search space, where the step size varies according to a normal distribution and improves the local best solution~\cite{Pearson1905problem}. The former approximates the global best solution according to satisfying the acceptance criterion given by the probability of the loudness $\mathcal{U}(0,1)<A^{(t)}_i$ (line 6 in Algorithm~\ref{alg:bat}), while the latter improves the local best solution according to the probability of pulse rate $r_i$ as:
\begin{equation}
    \mathbf{x}_i=\mathbf{x}_{best}+\mathcal{N}(0,1)*\bar{r},
    \label{eq:LRW}
\end{equation}
where the function $\mathcal{N}(0,1)$ denotes the random number drawn from the Gaussian distribution with mean zero and standard deviation one, and $\bar{r}$ designates the average value of the variable pulse rates, as found by all the individuals in the virtual swarm, i.e.~$\bar{r}=\frac{1}{Np}\cdot\sum_{i=1}^{Np} r_i$. Using two equations in the \texttt{BA} algorithm (i.e., Eq.~\ref{eq:bat} and Eq.~\ref{eq:LRW}) enables the user to balance exploration/exploitation processes of the evolutionary search explicitly~\cite{crepinsek2013exploration}.

The pseudo-code of the \texttt{BA} is illustrated in Algorithm~\ref{alg:bat}, where 
\begin{algorithm}[htb]
\caption{Original Bat algorithm}
\label{alg:bat}
\begin{algorithmic}[1]
\Procedure{bat\_algorithm}{}
\Require loudness reduction $A$, pulse rate increase $\gamma$, frequencies $Q_{\min}$ and $Q_{\max}$
\State \textsc{Initialize}($P^{(0)},\forall i:fitness_i=0,A_i=A,r_i=\gamma$); \Comment {initialization}
\While {$\mathbf{not}$ \textsc{TerminationConditionMeet}}
\ForAll {$i\in[1,\mathit{Np}]$}
\State $f_{trial}$=\textsc{Evaluate}$(\mathbf{x}_i)$; \Comment {evaluation of population}
\If {$f_{trial} > fitness_i~\mathbf{and}~\mathcal{U}(0,1) < A_i$} \Comment{acceptance criteria}
\State $fitness_i=f_{trial}; A_i=A_i*A; r_i=r_i*(1-\exp^{-\gamma*(t+1)})$;
\EndIf
\If{$f_{trial}>f_{best}$} \Comment{update the global best if needed}
\State $\mathbf{x}_{best}=\mathbf{x}_i;~f_{best}=f_{trial};$
\EndIf
\State $\mathbf{x}_i=$\textsc{Move}$(\mathbf{x}_i,\mathbf{x}_{best},Q_{\min},Q_{\max})$ \Comment{move $i$-th bat according Eq.~(\ref{eq:bat})}
\If{$\mathcal{U}(0,1)>r_i$} \Comment{Gaussian random walk}
\State $\mathbf{x}_i=\mathbf{x}_{best}+\mathcal{N}(0,1)*\bar{r}$;
\EndIf
\EndFor	
\EndWhile
\EndProcedure
\end{algorithmic}
\end{algorithm}
the function \textsc{Initialize} performs a random initialization of the particles with elements drawn from uniform distribution in the interval $[0,1]$ (denoted as $\mathcal{U}(0,1)$). The fitness vector elements are initialized to zero, while the adaptive loudness reduction $A^{(t)}_i$ and the pulse rate coefficient $r^{(t)}_i$ are initialized with either $A$ or $\gamma$ respectively. Thus, the \texttt{SA} component is implemented by lines 6-8, and the \texttt{GRW} by lines 13-15 in Algorithm~\ref{alg:bat}.

\section{Framework for identifying the strong equivalence of NI algorithms}\label{sec:3}

\begin{figure}[htb]
    \centering
    \includegraphics[width=0.92\linewidth]{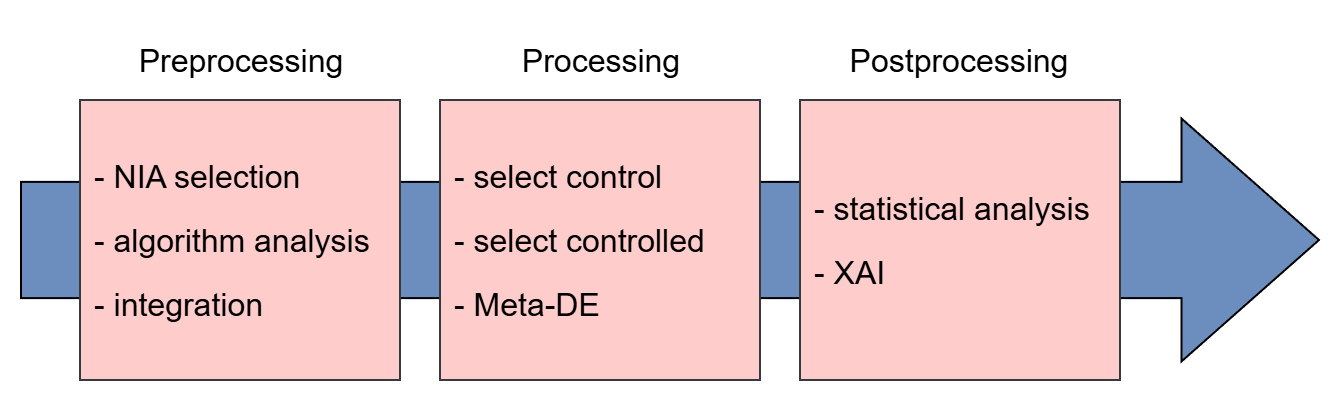}
    \caption{Workflow of the framework for identifying strong equivalence between NI metaheuristics.}
    \label{fig:flow}
\end{figure}

The aim of the framework for identifying the strong equivalence of NI metaheuristic algorithms (Fig.~\ref{fig:flow}) is to record the behavior of the reference NI algorithm (also control) by solving an arbitrary problem, and, then, by tuning the hyper-parameters of the referenced algorithm (also controlled), searches for those behaviors aggregated in the corresponding feature vectors that matches the reference algorithm as closely as possible according to the similarity metrics. The tuning process is performed automatically using the metaheuristic control system. The behavior of any algorithm during a single independent run is described by various descriptive metrics, whose values are accumulated into a feature vector. The strong equivalence is identified by comparing the feature vector of the control algorithm with the feature vector of the controlled one according to the similarity metrics. The best feature vector identified through the comparison is finally the subject of a detailed statistical analysis and eXplainable Artificial Intelligence (XAI)~\cite{Barredo2020explainable} methods.

As is evident from Fig.~\ref{fig:flow}, the proposed framework is outlined as a workflow consisting of three steps:
\begin{itemize}
\item preprocessing,
\item processing,
\item postprocessing.
\end{itemize}
The first step captures all the tasks necessary for integration of the NI algorithms into the framework. Mainly, the processing step encompasses an analysis of the implementation of the Meta-DE control system. Finally, the postprocessing includes those tasks that are suitable for the further results' processing.

\subsection{Preprocessing in framework}
The preprocessing step demands the following three tasks to be accomplished before identifying the strong equivalence between NI algorithms is conducted:
\begin{itemize}
    \item selecting an $\mathit{NIA}$ set consisting of the NI algorithms entering the framework,
    \item deep analysis of implementations of all the algorithms collected in the $\mathit{NIA}$ set,
    \item integration of algorithms from the $\mathit{NIA}$ set into the framework.
\end{itemize}
The mentioned tasks are discussed in detail in the remainder of the paper, 

\subsubsection{Selecting a set of the NI algorithms}
Selecting the set of NI algorithms $\mathit{NIA}$ means that a software engineer determines those algorithms for which the strong equivalence must be identified among the arbitrary selected pairs of algorithms from the set. Let the set of NI algorithms is given as:
\begin{equation}
    \mathit{NIA}=\{Alg_1,\ldots,Alg_Q\},
\end{equation}
where each element $Alg_i$ for $i=1,\ldots,Q$ denotes the particular NI algorithm from an $\mathit{NIA}$ set, and variable $Q$ designates the size of the set. 

\subsubsection{Deep analysis of implementations of the algorithms in the \textit{NIA} set}\label{sec:deep}
Deep analysis of implementations of the NI algorithms in \textit{NIA} is important for using the framework. During the analysis, a software engineer needs to become aware of how particular NI algorithm operate, i.e., which specific components incorporate, how many, and how do the parameters affect the result quality, and how similar are the definitions of the variation operators. This knowledge is crucial for deciding if the  NI algorithms in the \textit{NIA} can be put on a common denominator for identifying the strong equivalence.

Algorithm analysis is devoted for establishing the differences in implementations of the NI algorithms from three points of view:
\begin{itemize}
    \item component,
    \item parameter setting,
    \item variation operator.
\end{itemize}
In the remainder of the paper, the mentioned analyses are described in detail.

\paragraph{Component analysis}
Typically, each NI algorithm consists of the following seven components: 
\begin{itemize}
    \item representation, 
    \item initialization, 
    \item fitness function evaluation, 
    \item variation operators, 
    \item survivor selection, 
    \item termination condition, 
    \item optional adaptation and hybridization. 
\end{itemize} 
Although the EAs support more kinds of representation of individuals (e.g., binary strings, real-valued vectors, finite-state automata, tree structures in LISP), mainly, the metaphor based metaheuristic algorithms support population members represented as real-valued vectors. Typically, the randomly generated individuals are placed in the initial population by the NI algorithms. However, we must ensure the same starting conditions in respect of the initial population for all the observed NI algorithms. The fitness function evaluation depends on the particular problem that needs to be solved by the definite NI algorithm, and, therefore, it is independent of the framework. The variation operator in metaphor based algorithms is usually the 'move' operator, that creates the new individuals from the old ones based on the definite formula, which describes the most detailed applied metaphor. Although a lot of survivor selection (also replacement) schemes can be found in EC, we don't have such comfort by SI-based algorithms, because, normally, these algorithms either don't support the survivor selection operator at hand, or apply a one-to-one replacement operator, respectively. This operator is also connected closely with the population model used. For instance, the no-replacement operator demands 1-population model, where each new created individual in the population survives and transfers its genetic characteristics into the new generation, while the one-to-one selection demands a 2-population model, where only the better between parents and its offspring will survive in the next generation. Although there are several ways for how to terminate the NI-algorithm, usually, the predefined maximum number of generations is employed. Optionally, the NI algorithms can improve their behavior by incorporating problem knowledge in the form of hybridization and adaptation.

\paragraph{Parameter setting analysis}
The parameters of NI algorithms are crucial for their good performance~\cite{eiben2015introduction}. However, various NI algorithms support different numbers of algorithm parameters, with which the behavior is controlled of the NI metaheuristics. With this in mind, the proposed framework needs to find the behavior of controlled $Alg2$ using $n_2$ parameters that is the most equivalent the control $Alg1$ using $n_1$ parameters in the sense of the equivalence theorem. 

In general, we distinguish two kinds of altering parameter values of NI algorithms:
\begin{itemize}
    \item tuning,
    \item parameter control.
\end{itemize}
The tuning process is applied by the metaheuristic control system for searching the optimal initial values of definite algorithm parameters before the evolutionary run. On the other hand, parameter control is implemented by the observed NI algorithm for altering their values during the evolutionary run implicitly. Normally, the NI community distinguishes the following four categories of parameter settings by the NI algorithms:
\begin{itemize}
    \item fixed,
    \item deterministic,
    \item adaptive,
    \item self-adaptive.
\end{itemize}
The fixed parameter control does not alter any parameter values, i.e., the parameter values stay unaltered. The deterministic parameter control means that the parameter values are altered according to some deterministic rule. The parameter values change regarding the feedback from the search process in adaptive parameter control, while these values undergo acting the variation operators, together with the corresponding problem variables, by self-adaptive parameter control.

At first, the result of the parameter setting analysis is to identify especially those algorithm parameters that affect the behavior of the variation operator. Thus, their optimal values can be determined during the extensive tuning process. Then, the parameter control needs to be established for each of these algorithm parameters. Finally, a domain of feasible values must be specified for the algorithm parameters at hand. 

\paragraph{Variation operators analysis}
The algorithm parameters control the behavior of several components of the NI algorithms. For instance, the parameter $\mathit{Np}$ indicates the maximum number of individuals within a population. Such control parameters are common for the majority of the NI algorithms, and, therefore, their settings are not so delicate when the same values are set by all the NI algorithms in the proposed framework. More interesting are the algorithm parameters for controlling the behavior of the variation operators because their setting can affect the results of identifying the equivalence of NI algorithms critically.

In SI-based algorithms, the variation operator consists primarily of a base vector to which one or two terms are added. Each additive term is represented as a difference of two vectors taken from the population of individuals and scaled by a scale factor, which can be set randomly or dependently on the used metaphor. The result of the analysis is illustrated in syntax diagrams or expressions in a simplified BNF notation. Both forms are suitable for establishing the similarities and differences between the observed variation operators.

\subsubsection{Integration of NI algorithms from \textit{NIA} into the framework}
The purpose of the section is to describe how to prepare each algorithm in an \textit{NIA} set to conform to the demands of the framework. At first, an implementation of a particular NI algorithm may be found on Git or other web sources. It is important that all the NI algorithms in the \textit{NIA} are implemented in the same programming language. Thus, the language specific implementation differences can be eliminated and, therefore, the comparisons between the NI algorithms can be conducted more fairly. 

The integration of the implemented NI algorithms into the framework takes place in three steps, as follow: 
\begin{itemize}
    \item recording the behavior of the NI algorithm during the evolutionary run,
    \item definition of the feature vector,
    \item tuning the behavior of the NI algorithms by using the passed parameters.
\end{itemize}
Recording the behavior of the NI algorithm demands the inclusion of an additional code for calculating the appropriate metrics. These metrics are aggregated into a feature vector suitable for further comparative analysis. Passing parameters represent an operating system's ability for controlling the behavior of the launched program, either automatically by metaheuristic control system, or manually by a software engineer, respectively.

\paragraph{Recording the behavior of the NI algorithm}
The proposed framework records the behavior of the NI algorithm using the various descriptive metrics after each generation. The metrics are measured in both the phenotype, as well as the genotype space. Two types of descriptive metrics are distinguished, as follow:
\begin{enumerate}
    \item population,
    \item individual.
\end{enumerate}

The population descriptive metrics are focused on aggregation of the properties of population members in any specific generation, while the individual metrics summarize the behavior of any particular individual across all the generations. 

The mentioned study considers the following four population metrics:
\begin{enumerate}
    \item fitness-distance correlation,
    \item diversity of the population,
    \item population fitness mean,
    \item population fitness standard deviation,
\end{enumerate}
and the following four individual metrics:
\begin{enumerate}
    \item individual distance traveled,
    \item individual fitness inter-quartile range,
    \item individual fitness mean,
    \item individual sinuosity index.
\end{enumerate}
In the remainder of the paper, the mentioned descriptive metrics are discussed in detail.

\textbf{Fitness-distance correlation}. Given a set $F^{(t)}=\{f^{(t)}_1,f^{(t)}_2,\ldots,f^{(t)}_{\mathit{Np}}\}$ of $\mathit{Np}$ individual fitness values and a corresponding set $D^{(t)}=\{d^{(t)}_1,d^{(t)}_2,\ldots,d^{(t)}_{\mathit{Np}}\}$ of $\mathit{Np}$ distances to the nearest global maximum in generation $t$, we compute the correlation coefficient $r^{(t)}$ as~\cite{Jones1995fitness}:
\begin{equation}
    r^{(t)}=\frac{c^{(t)}_{FD}}{s^{(t)}_F\cdot s^{(t)}_D},
\end{equation}
where
\begin{equation}
    c^{(t)}_{FD}=\frac{1}{\mathit{Np}}\sum_{i=1}^{\mathit{Np}} (f^{(t)}_i-\bar{f}^{(t)})\cdot(d^{(t)}_i-\bar{d}^{(t)}),
\end{equation}
is the covariance of $F^{(t)}$ and $D^{(t)}$, $s^{(t)}_F$ and $s^{(t)}_D$ are the standard deviations, and $\bar{f}^{(t)}$ and $\bar{d}^{(t)}$ are the means, and $\mathit{Np}$ is the population size.

\textbf{Diversity of the population}. According to \citet{Ursem2002diversity}, the diversity of a population is defined mathematically as follows:
\begin{equation}
    diversity(P^{(t)})=\frac{1}{|L|\cdot \mathit{Np}}\sum_{i=1}^{\mathit{Np}}\sqrt{\sum_{j=1}^D (x^{(t)}_{i,j}-\bar{x}^{(t)}_j)},
\end{equation}
where $|L|$ is the length of the diagonal in the search space $S\subseteq R^{D}$, $P^{(t)}$ is the population in generation $t$, $\mathit{Np}$ is the population size, $D$ is the dimensionality of the problem, $x^{(t)}_{i,j}$ is the $j$-th value of the $i$-th individual, and $\bar{x}^{(t)}_j$ is the mean value of the j-th solution element in the population (i.e., the centroid). 

\textbf{Population fitness mean}. Given a set $F^{(t)}=\{f^{(t)}_1,f^{(t)}_2,\ldots,f^{(t)}_{\mathit{Np}}\}$ of $\mathit{Np}$ individual fitness values in generation $t$, the population fitness mean $\bar{f}$ is defined as follows:
\begin{equation}
    \bar{f}^{(t)}=\frac{1}{\mathit{Np}}\sum_{i=1}^{\mathit{Np}} f^{(t)}_i.
\end{equation}

\textbf{Population fitness standard deviation}. Given a set $F^{(t)}=\{f^{(t)}_1,f^{(t)}_2,\ldots,f^{(t)}_{\mathit{Np}}\}$ of $\mathit{Np}$ individual fitness values in generation $t$, the population fitness standard deviation $s^{(t)}_F$ is defined as follows:
\begin{equation}
    s^{(t)}_F=\sqrt{\frac{\sum_{i=1}^N (f^{(t)}_i-\bar{f}^{(t)})^2}{\mathit{Np}}},
\end{equation}
where $\bar{f}^{(t)}$ is the population fitness mean.

\textbf{Individual distance traveled}. Individual Distance Traveled (IDT) measures the length of the search space trajectories traveled by all the individuals $\langle \mathbf{x}^{(t)}_i,\mathbf{x}^{(t+1)}_{i}\rangle$, in other words: 
\begin{equation}
    \mathit{IDT}_i=\sum_{t=1}^{T-1}d(\textbf{x}^{(t)}_{i},\textbf{x}^{(t+1)}_{i}),
\end{equation}
Here, $d(\textbf{x}^{(t)}_{i},\textbf{x}^{(t+1)}_{i})$ denotes the Euclidean distance expressed as:
\begin{equation}
    d(\textbf{x}^{(0)}_i,\textbf{x}^{(T)}_i)=\sqrt{(\textbf{x}^{(0)}_i-\textbf{x}^{(T)}_i)^2}.    
\end{equation}

\textbf{Individual Fitness Inter-Quartile Range}. The Individual Fitness Inter-Quartile Range (IFIQR) is defined as follows: Given a set $F_i=\{f^{0}_i,f^{1}_i,\ldots,f^{(T)}_i\}$ for $i=1,\ldots,\mathit{Np}$ individual fitness ordered ascending as $F'_i=\{f'^{(0)}_i,f'^{(1)}_i,\ldots,f'^{(T)}_i\}$ 
where the IFIQE is the difference between the third quartile $Q3$ and the first quartile $Q1$, more precisely:
\begin{equation}
    \mathit{IFIQR}_i=Q3_i-Q1_i.
\end{equation}

\textbf{Individual Fitness Mean}. Given a set $F_i=\{f^{0}_i,f^{1}_i,\ldots,f^{(T)}_i\}$ for $i=1,\ldots,\mathit{Np}$, the Individual Fitness Mean (IFM) is defined mathematically as:
\begin{equation}
    \mathit{IFM}_i=\frac{1}{T}\sum^{T}_{t=0}f^{(t)_i}.
\end{equation}

\textbf{Individual Sinuosity Index}. The Individual Sinuosity Index (ISI) is the ratio of the curvilinear length along the curve (IDT) and the Euclidean distance (straight line) between the end points of the curve. The mathematical definition of the measure is straightforward: Given a starting population $\textbf{x}^{(0)}_i\in P^{(0)}$ and final population $\textbf{x}^{(T)}_i\in P^{(T)}$ for $i=1,\ldots,\mathit{Np}$, the ISI is defined as:
\begin{equation}
    \mathit{ISI}_i=\frac{\mathit{IDT}_i}{d(\textbf{x}^{(0)}_i,\textbf{x}^{(T)}_i)},
\end{equation}
where $d(x^{(0)}_i,x^{(T)}_i)$ is a Euclidean distance. 

\paragraph{Definition of the feature vector}
The feature vector is assembled consisting of the population and individual descriptive metrics, as follows:
\begin{equation}
    \mathbf{FV}^{(Alg)}=\langle \underbrace{r^{(t)},diversity(P^{(t)}),\bar{f}^{(t)},s^{(t)}_F}_{\text{for}~t=1,\ldots,T},\underbrace{\mathit{IDT}_i,\mathit{IFIQR}_i,\mathit{IFM}_i,\mathit{ISI}_i}_{\text{for}~i=1,\ldots,\mathit{Np}}\rangle,
    \label{eq:classifier}
\end{equation}
where $(Alg)$ denotes the NI metaheuristic at hand, and the first under-brace designates the particular population descriptive metrics, while the second one the particular individual descriptive metrics. As is evident from Eq.~(\ref{eq:classifier}), the length of the feature vector $\mathbf{FV}^{(Alg)}$ is expressed as $|\mathbf{FV}^{(Alg)}|=4\times (T+\mathit{Np})$. The feature vector is a result of the optimization by both, i.e., the control and controlled algorithms, and saved into a predefined directory on the disk.

\paragraph{Passing parameters}
The program (typically the name of the NI algorithm) is called within the operating system shell. It is controlled using switches and arguments. The former affect how the program is interpreted, while the latter denote the objects on which it is executed. The framework supports the switches, as illustrated in Table~\ref{tab:passing}, 
\begin{table}[htb]
    \caption{Passing parameter.}
    \label{tab:passing}
    \centering
    \begin{tabular}{c|l}
        \hline
         Switch &  Argument \\
        \hline\hline
         $-\text{a}$ to $-\text{c}$  & Various algorithm's parameters \\
         $-\text{g}$ & Maximum number of generations \\
         $-\text{d}$ & Problem dimension \\
         $-\text{n}$ & Number of parameters \\
         $-\text{r}$ & Maximum number of independent runs \\
         $-\text{s}$ & Random generator seed \\
         $-\text{f}$ & Path to the output directory \\
         \hline
    \end{tabular}
\end{table}
from which it can be seen that the switches '-a', '-b', and '-c' are employed for passing the algorithm's parameters on the called program. The other switches are self-explainable, and, therefore, do not need any additionally explanation. 

Let us emphasize that the number of switches is not fixed and can be extended by a software engineer if necessary. On the other hand, the switch and argument are passed to the called program without an intermediate blank character.

\subsection{Processing in framework}
Processing represents the core of the framework and covers two main tasks, as follow:
\begin{itemize}
    \item control/controlled algorithm selection,
    \item design and implementation of Meta-DE.
\end{itemize}
In the remainder of the paper, the mentioned tasks are discussed in detail.

\subsubsection{Control/controlled algorithm selection}
The aim of the task is to determine the control algorithm from the $\mathit{NIA}$ set. After the selection, all the other NI algorithms in the $\mathit{NIA}$ set become controlled, The control algorithm is responsible for creating the reference behavior, to which the controlled algorithms try to adapt their behaviors as closely as possible by tuning their hyper-parameters. Indeed, the framework is able to identify the strong equivalence between an arbitrary pair of NI algorithms in the $\mathit{NIA}$. In accordance with this, the arbitrary pair of algorithms is expressed as $(Alg1,Alg2)$, where $Alg1\in\mathit{NIA}$ and $Alg2\in\mathit{NIA}$. For the framework it is also valid that the $Alg2$ is $Alg1$, in other words $Alg1=Alg2$.

\subsubsection{Design and implementation of \texttt{Meta-DE}}

\begin{figure}[htb]
    \centering
    \includegraphics[width=0.8\linewidth]{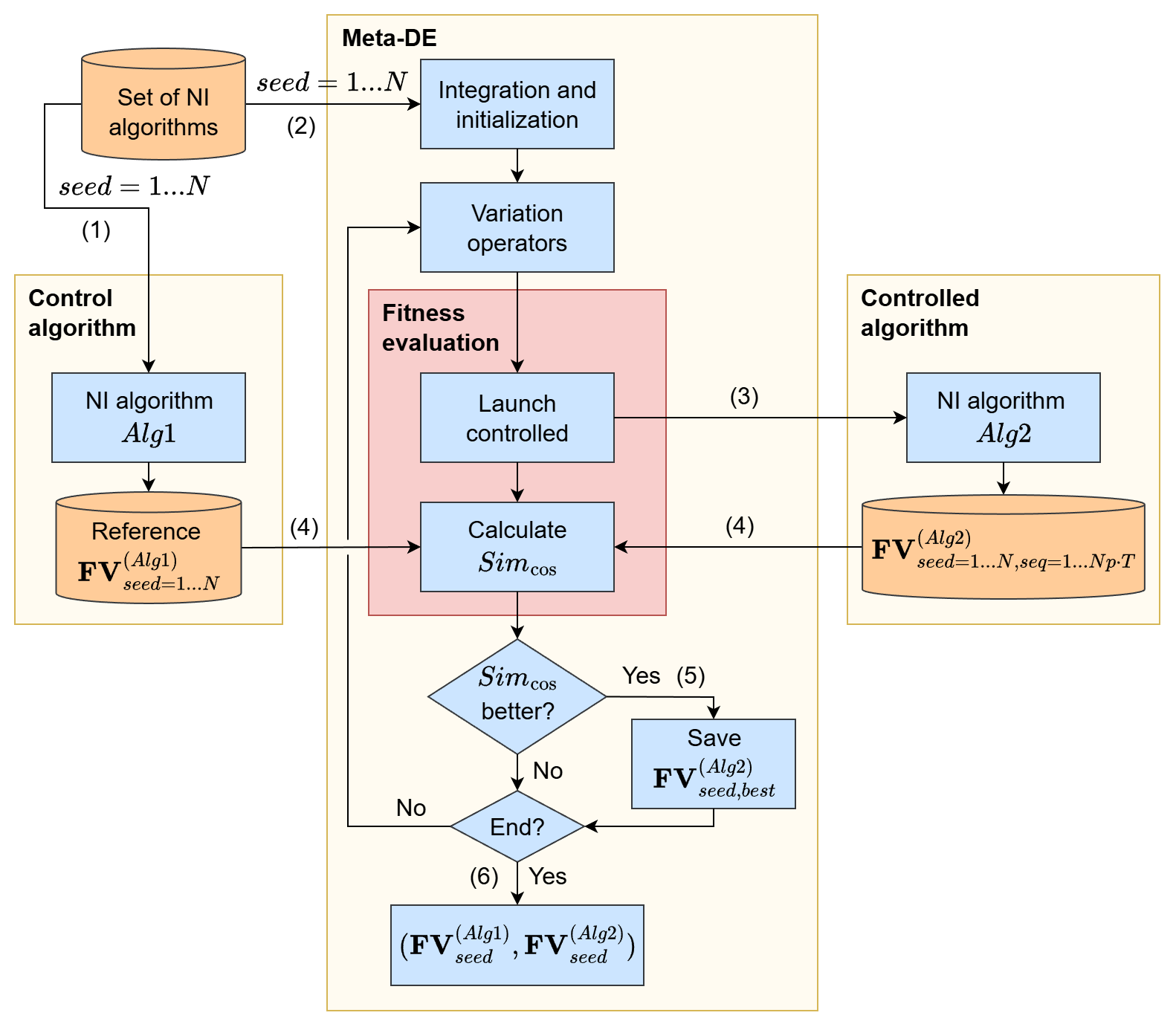}
    \caption{Scheme of the framework for identifying the similarity of NI algorithms, where $Alg1$ denotes the control and $Alg2$ the controlled algorithm.}
    \label{fig:framework}
\end{figure}

The scheme of the framework for identifying the strong equivalence of the NI metaheuristic algorithms is depicted in Fig.~\ref{fig:framework}, 
from which it can be observed that the framework consists of three components:
\begin{itemize}
    \item control algorithm $Alg1$,
    \item controlled algorithm $Alg2$,
    \item metaheuristic control system \texttt{Meta-DE}.
\end{itemize}
While the pair of control and controlled algorithms is a simple implementation of corresponding NI metaheuristics integrated into the framework, the \texttt{Meta-DE} algorithm is a metaheuristic control system for the purpose of controlling the hyper-parameters of the controlled algorithm. Although the idea behind the metaheuristic control system proposed by \citet{Grefenstette1986optimization} is not new, the concept has still been employed commonly in the NI community. 

In the proposed framework the original \texttt{DE} algorithm, as proposed by \citet{storn1997differential}, was selected for developing of \texttt{Meta-DE} control system, due to implementation simplicity and expected good results. The proposed \texttt{Meta-DE} operates with a population of solutions, where each solution is represented as a real-valued vector whose elements denote a value of one hyperparameter of the controlled algorithm. Mathematically, each individual is expressed in \texttt{Meta-DE} as:
\begin{equation}
    \mathbf{x}^{(t)}_i=\left( \underbrace{x^{(t)}_{i,1}}_{parm_1},\ldots,\underbrace{x^{(t)}_{i,Q}}_{parm_Q}\right),
\end{equation}
where $x^{(t)}_{i,j}\mapsto parm_j$ designates a genotype/phenotype mapping of a definite element of vector $x^{(t)}_{i,j}\in[lb^{(t)}_j,ub^{(t)}_j]$ in the genotype to the value of the corresponding algorithm's parameter $parm\_j$ in the phenotype space, the variable $Q$ denotes the number of controlled hyper-parameters, and the $lb^{(t)}_j$ and $ub^{(t)}_j$ are the boundaries of the interval of the feasible values for the hyper-parameter. 

In fact, the \texttt{Meta-DE} is the classical EA. In line with this, it incorporates the components that are illustrated within the flowchart \texttt{Meta-DE} box in Fig.~\ref{fig:architecture} as follows: (1) initialization ('Integration and initialization' box), (2) \texttt{DE} mutation strategy ('Variation operators' box), (3) fitness function evaluation ('Fitness evaluation' box), (4) survivor selection ('$Sim_{\cos}$ better?' decision box), and (5) termination condition ('End?' decision box). The Meta-DE employs: random initialization of the population, standard 'DE/rand/1/bin' mutation strategy for the variation of vectors, one-to-one selection for determining those vectors that will survive in the next generation, while the maximum number of generations is used as a termination condition. The fitness function evaluates the equivalence of two NI algorithms based on the cosine similarity of their feature vectors. The cosine similarity is explained in the remainder of the paper in more detail. The final result of the \texttt{Meta-DE} processing is the feature vector $\mathbf{FV}^{(Alg2)}_{seed,best}$ that exhibits the nearest behavior of the controlled algorithm $Alg2$ to the control $Alg1$ according to the cosine similarity metric.

\paragraph{Framework in action}
Let us emphasize that the framework operates as a workflow consisting of a sequence of steps, where each step is denoted by the corresponding sequence numbers within the round brackets, as depicted in Fig.~\ref{fig:framework}:
\begin{description}
    \item[\textrm{(1)}] At first, the control algorithm $Alg1$ needs to be selected from an \textit{NIA} set of NI algorithms that are already integrated into the framework for further analysis. One of the $Alg_i\in NIA$ is selected for the control algorithm $Alg1$ by the framework user. The control algorithm generates $N$ different reference feature vectors $\mathbf{FV}^{(Alg1)}_{seed}$, where each independent run starts with a different seed of random number generator $seed=1,\ldots,N$, and $N$ designates the maximum number of independent runs of particular NI metaheuristics. The results of the optimization is a collection of feature vectors saved in files under the names "Alg1.seed".
    \item[(2)] For all the other $Alg_i\in\mathit{NIA}$, playing a role of controlled algorithms, only one must be selected as the controlled algorithm $Alg2$ in each \texttt{Meta-DE} run. Before initialization, the Meta-DE adapts the length of the genotype vector properly. Typically, the number of independent runs corresponds to the number of different seeds of random number generator $seed=1,\ldots,N$. The same seeds ensure the same initial conditions (i.e., the same initial population) by comparing the algorithms $Alg1$ and $Alg2$. In the case that the $Alg2$ is $Alg1$, the \texttt{Meta-DE} tries to reconstruct the original values of the $Alg1$ control parameters by tuning the randomly initialized hyper-parameters of the $Alg2$. 
    \item[(3)] The heart of the \texttt{Meta-DE} control system represents the fitness function evaluation that differs from the classical black-box optimization of NI algorithms. Actually, the analytical function for evaluating the quality of the solution is replaced by searching for the evaluation to reach some expected behavior the best~\cite{Doncieux2014beyond}. However, the expected behavior is not evaluated inside the \texttt{Meta-DE} control system, but by launching the external NI program implementing $Alg2$, and controlling with the corresponding hyper-parameters passed by the operating system feature. Let us mention that the behavior of the controlled $Alg2$ is saved into the predefined directory on the disk, to which also \texttt{Meta-DE} has access. In summary, the $Alg2$ is launched $\mathit{Np}\cdot T$ times. Thereby, the same number of the feature vectors $\mathbf{FV}^{(Alg2)}_{seed,seq}$ that are saved under the names 'Alg2.\textit{seq}', where $seq=1,\ldots,\mathit{Np}\cdot T$.
    \item[(4)] The step is devoted for calculation of the cosine similarity between the feature vectors obtained by the control algorithm behavior $\mathbf{FV}^{(Alg1)}_{seed}$ and the one produced by the controlled algorithm $\mathbf{FV}^{(Alg2)}_{seed,seq}$. The result of the calculation is the analytical fitness value of the similarity $Sim_{\cos}(\mathbf{FV}^{(Alg1)}_{seed},\mathbf{FV}^{(Alg2)}_{seed,seq})$.
    \item[(5)] The step is executed only conditionally: in the case that the calculated value of $Sim_{\cos}$ from step~(4) is better that the existing one, the new best feature vector $\mathbf{FV}^{(Alg2)}_{seed,best}$ becomes $\mathbf{FV}^{(Alg2)}_{seed,seq}$.
    \item[(6)] After the termination condition is satisfied, the final result in the form $(\mathbf{FV}^{(Alg1)}_{seed},\mathbf{FV}^{(Alg2)}_{seed})$, is obtained, where $\mathbf{FV}^{(Alg2)}_{seed}=\mathbf{FV}^{(Alg2)}_{seed,best}$. 
\end{description}

This pair of control and controlled NI algorithms represents those couples that are the most similar in the sense of the equivalence theorem as found by the proposed framework. In order to analyze the equivalence even further, the postprocessing step is developed and explained in more detail in the remainder of the paper.

\subsection{Postprocessing in framework}
This section illustrates of the following three issues needed for understanding the foundations of the proposed framework: 
\begin{itemize}
    \item statistical analysis,
    \item visualization and XAI.
\end{itemize}
In the remainder of the paper, the mentioned issues are discussed in detail. 

\subsubsection{Statistical analysis}
The statistical analysis is founded on the introduction of various similarity metrics that are devoted for identifying additionally the strong equivalence of two NI metaheuristics and are based on a comparison of the feature vectors. In the present study, the following similarity metrics are defined for identifying the strong equivalence between two NI metaheuristic algorithms based on the similarity of feature vectors:
\begin{itemize}
\item cosine similarity ($Sim_{\cos}$),
\item Symmetric Mean Absolute Percentage Error ($Sim_{\text{SMAPE}}$),
\item Spearman's rank correlation coefficient ($\rho$),
\item classification accuracy of k-Nearest Neighbor ($Sim_{\text{kNN}}$),
\item classification accuracy of Support Vector Machine ($Sim_{\text{SVM}}$),
\item classification accuracy of Random Forrest ($Sim_{\text{RF}}$).
\end{itemize}
The similarity metrics used in the study can be divided into statistical (the first three) and ML classification ones (the last three). While the statistical similarity metrics are discussed in more detail in the remainder of the paper, the ML similarity classification metrics are general enough that they do not require any additional explanation. Let us emphasize that the ML similarity metrics are calculated from the classification accuracies of the corresponding ML methods according to the following equation:
\begin{equation}
    Sim_{(.)} = 1- 2*\text{abs}(0.5-Accuracy_{(.)}),
\end{equation}
where notation $(.)$ denotes an arbitrary selected ML method from a set $\{\text{kNN,SVM,RF}\}$, and the $Accuracy$ is referred to the particular classification accuracy.

\textbf{Cosine similarity}. The cosine similarity of two nonzero feature vectors $\mathbf{FV}^{(Alg_1)}$ and $\mathbf{FV}^{(Alg_2)}$is defined as:
\begin{small}
\begin{equation}
    Sim\left(\mathbf{FV}^{(Alg_1)},\mathbf{FV}^{(Alg_2)}\right)=\frac{\left|\mathbf{FV}^{(Alg_1)}\cdot \mathbf{FV}^{(Alg_2)}\right|}{\left|\left|\mathbf{FV}^{(Alg_1)}\right|\right|\cdot\left|\left|\mathbf{FV}^{(Alg_2)}\right|\right|}.
\end{equation}
\end{small}

\textbf{Symmetric mean absolute percentage error}. The Symmetric Mean Absolute Percentage Error $sim_{\text{SMAPE}}$ is an accuracy metric based on percentage (or relative) errors. It is defined as:
\begin{equation}
    Sim_{\text{SMAPE}}\left(\mathbf{FV}^{(Alg_1)},\mathbf{FV}^{(Alg_2)}\right)=1-\frac{1}{|\mathbf{FV}^{(.)}|}\mathlarger{\sum}^{|\mathbf{FV}^{(.)}|}_{i=1}\frac{\left|c^{(Alg_1)}_i-c^{(Alg_2)}_i\right|}{\left|c^{(Alg_1)}_i+c^{(Alg_2)}_i\right|},
\end{equation}
where $\mathbf{FV}^{(Alg_1)}$ and $\mathbf{FV}^{(Alg_2)}$ are the particular feature vectors at hand, and $|\mathbf{FV}^{(.)}|$ is their corresponding sizes. Let us mention that the metric is minimized, and needs to be subtracted from~1 when the observed problem is a maximization one. 

\textbf{Spearman's rank correlation coefficient}. Spearman's rank correlation coefficient $\rho\in[-1,1]$ indicates how strongly two sets of ranks obtained by two feature vectors $\mathbf{FV}^{(Alg_i)}=\{c^{(Alg_i)}_j\}$ for $i\in[1,2]$ and $j=1,\ldots |\mathbf{FV}|$ are correlated. Mathematically, it is defined as:
\begin{small}
\begin{equation}
    \rho\left(\text{R}[\mathbf{FV}^{(Alg_1)}],\text{R}[\mathbf{FV}^{(Alg_2)}]\right)=\frac{cov[\text{R}[\mathbf{FV}^{(Alg_1)}],\text{R}[\mathbf{FV}^{(Alg_2)}]]}{\sigma_{\text{R}[\mathbf{FV}^{(Alg_1)}]}\cdot \sigma_{\text{R}[\mathbf{FV}^{(Alg_2)}]}},
\end{equation}
\end{small}
where $cov[\text{R}[\mathbf{FV}^{(Alg_1)}],\text{R}[\mathbf{FV}^{(Alg_2)}]]$ is a covariance of ranks R[.] and $|\mathbf{FV}^{(.)}|$ denotes the size of a definite feature vector.

\section{Experiments and results}\label{sec:4}
The goal of the experimental work was to show that the proposed framework can find the equivalence of the NI algorithms from theoretical, as well as experimental points of view. From the former point of view, the deep analysis exhibits the more similar implementation features of the algorithms in \textit{NIA}, while, from the latter, the strong equivalence of the NI algorithms is identified experimentally. On the one hand, the theoretical similarity can be very high for some NI algorithms, while, on the other, the experimental results of identifying the strong equivalence indicates the opposite. The test of the framework for identifying the equivalence of NI algorithms follows the steps as demanded by the framework's workflow. The detailed results of the tests are illustrated in detail in the remainder of the paper.

\subsection{Selecting a set of NI metaheuristics}

In the mentioned study, the proposed framework was applied to the following set of NI metaheuristic algorithms: (1) \texttt{accPSO}, (2) \texttt{accFA}, (3) \texttt{PSO}, (4) \texttt{FA}, and (5) \texttt{BA}. Mathematically, the \textit{NIA} set is expressed as:
\begin{equation}
    \mathit{NIA}=\{\texttt{accPSO,accFA,PSO,FA,BA}\},
\end{equation}
where each element of the set denotes a specific NI algorithm.

\subsection{Algorithm analysis}\label{sec:alg_anal}

\subsubsection{Component analysis of algorithms in \textit{NIA}}

The results of the deep analysis of the  implemented components in the NI algorithms from the \textit{NIA} are aggregated in Table~\ref{tab:components}, where the component characteristics of the mentioned algorithms are exposed by the used representation of individuals, variation operators, parent selection schemes, population models, and potential adaptation mechanisms. As can be observed from the table, 
\begin{table}[htb]
    \caption{Summary characteristics of the NI metaheuristics in the study.}
    \label{tab:components}
    \centering
    \begin{tabular}{l|lllll}
     \hline
    \multirow{2}{*}{Algorithm} & Representation & Variation  & Selection & Population & Adaptation \\
     & of individuals & operator & scheme & model & (optional) \\
     \hline
      \texttt{accPSO} & Floating-point & Eq.~(\ref{eq:apso}) & global best & 1-population & n/a \\
      \texttt{accFA}  & Floating-point & Eq.~(\ref{eq:afa}) & global best  & 1-population & n/a \\
      \texttt{PSO}    & Floating-point & Eq.~(\ref{eq:pso}) & one-to-one & 2-populations & n/a \\
      \texttt{FA}   & Floating-point & Eq.~(\ref{eq:fa}) & global best & 1-population &  n/a \\
      \texttt{BA}    & Floating-point & Eq.~(\ref{eq:bat}) & \texttt{SA}-selection\footnotemark[1] & 1-population & Eq.~(\ref{eq:LRW}), \texttt{GRW}\footnotemark[2]\\
      \hline
    \end{tabular}
\end{table}
\footnotetext[1]{\texttt{SA}-selection Excluded when $A_i=1$.}
\footnotetext[2]{\texttt{GRW} excluded when $r_i=0$.}
all the algorithms employed the same (i.e., floating-point ) representation of individuals, different variation operators whose implementations are inspired by their metaphor, different parent selection schemes (i.e., from 'global best', where no selection is taken because all solutions are attracted by their global best solution only, next to 'one-to-one' selection, where the best between the offspring and parent individual will survive in the next generation, and, finally, to 'SA' selection, where the the best solution is selected for the next generation conditionally with some positive probability).

\subsubsection{Parameter setting analysis of the algorithms in the \textit{NIA}}
The parameter setting analysis captures only those algorithm parameters that control the behavior of the variation operators directly. In this study three different aspects of their applicability can be indicated by the software engineer, as follow:
\begin{enumerate}
    \item setting the initial values of the parameters,
    \item altering the algorithm parameters by the parameter control mechanisms,
    \item determining the proper domains of feasible values for each algorithm parameter. 
\end{enumerate}
The results of the analysis according to the exposed aspects are represented in Table~\ref{tab:param_set}. As is evident from the table,
\begin{table}[htb]
    \caption{Parameter analysis of the observed NI metaheuristic algorithm.}
    \label{tab:param_set}
    \centering
    \begin{tabular}{l|lll|lll|lll}
    \hline
    \multirow{2}{*}{Algorithm} & \multicolumn{3}{c|}{Initial} & \multicolumn{3}{c|}{Parameter control} & \multicolumn{3}{c}{Domain} \\
    \cmidrule{2-4}\cmidrule{5-7}\cmidrule{8-10}
     & 1 & 2 & 3 & 1 & 2 & 3 & 1 & 2 & 3 \\
    \hline
     \texttt{accPSO} & $\alpha_0$ & $\beta_0$ & n/a & FIX & FIX & n/a & $[0.1,1.0]$ & $[0.1,1.0]$ & n/a \\
     \texttt{accFA} & $\alpha_0$ & $\beta_0$ & n/a & FIX & FIX & n/a & $[0.1,1.0]$ & $[0.1,1.0]$ & n/a \\
     \texttt{PSO} & $w$ & $c_1$ & $c_2$ & FIX & FIX & FIX & $[0.4,0.9]$ & $[1.5,2.5]$ & $[1.5,2.5]$ \\
     \texttt{FA} & $\alpha_0$ & $\beta_0$ & $\gamma$ & DET & META & FIX & $[0.1,1.0]$ & $[0.5,1.5]$ & $[0.1,1.0]$ \\
     \texttt{BA} & $A_0$ & $\gamma$ & n/a & DET & META & n/a & $[0.7,1.2]$ & $[0.1,0.3]$ & n/a\\
    \hline
    \end{tabular}
\end{table}
the \texttt{PSO} and \texttt{FA} algorithms support three, while the other (i.e., \texttt{accPSO}, \texttt{accFA}, and \texttt{BA}) only two parameters. These are initialized using the operating system passing parameters mechanisms, either by the software engineer manually or by the \texttt{Meta-DE} automatically. The initialized parameters remain unchanged during the whole evaluation run by algorithms \texttt{accPSO}, \texttt{accFA}, and \texttt{PSO}, as designated by the fixed (tag 'FIX' in the table) parameter control. The parameters $\alpha_0$ by the \texttt{FA} and $A_0$ by the \texttt{BA} algorithms are altered using a simple deterministic rule (tag 'DET' in the table), while the parameter $\beta$ by the \texttt{FA} and $\gamma$ by the \texttt{BA} algorithms are altered using sophisticated deterministic function (tag 'META' in the table) mimicking a metaphor from the nature. The domains of feasible values are important for the Meta-DE, which alters the initial values of the definite algorithm parameters within the prescribed domains by searching for the better strong equivalence.  

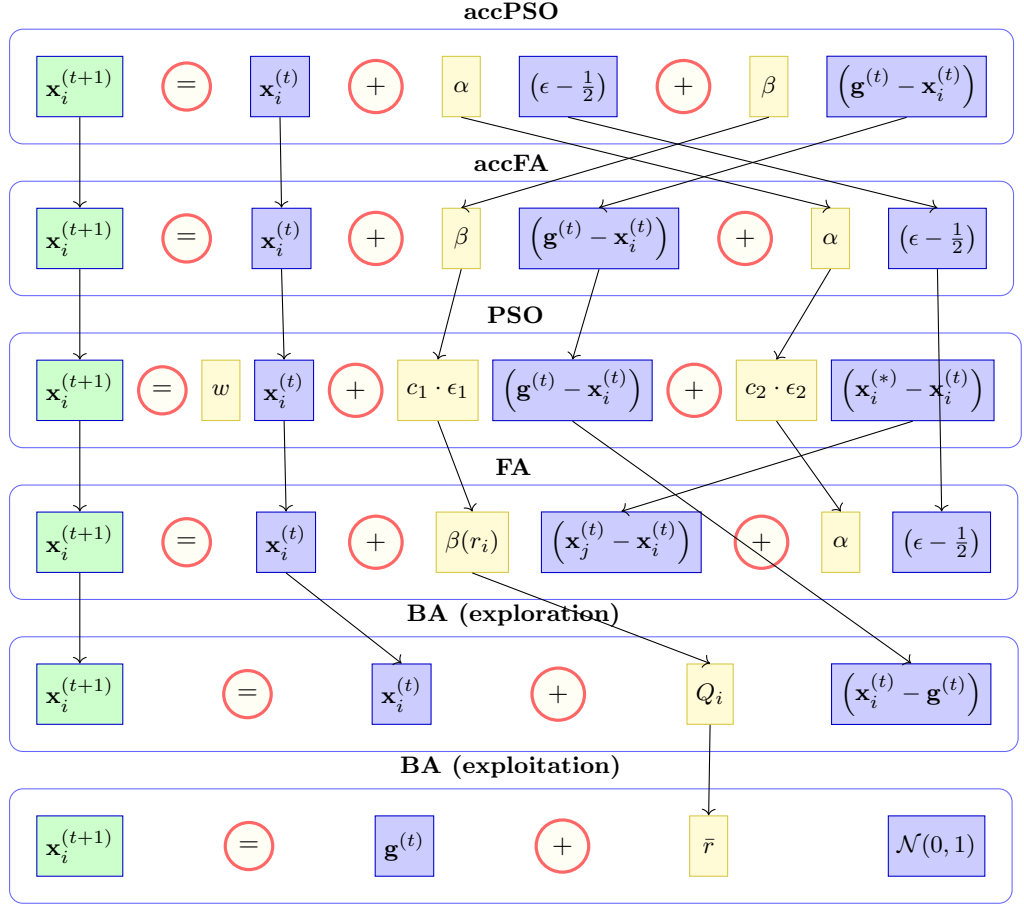
\begin{figure}[hbt]
\centering
\begin{tikzpicture}[
    square_node_blue/.style={
      draw=blue!80!black, rectangle, minimum width=0.5cm, minimum height=0.8cm,
      fill=blue!20, text=black, align=center, font=\small
    },
    square_node_green/.style={
      draw=blue!80!black, rectangle, minimum width=0.5cm, minimum height=0.8cm,
      fill=green!20, text=black, align=center, font=\small
    },
    square_node_yellow/.style={
      draw=yellow!80!black, rectangle, minimum width=0.5cm, minimum height=0.8cm,
      fill=yellow!20, text=black, align=center, font=\small
    },
    round_node/.style={circle, draw=red!60, fill=yellow!5, very thick, minimum size=5mm},
    box/.style = {draw,red,inner sep=10pt,rounded corners=5pt}] 
  ]

\node[square_node_green]       (target)          {$\mathbf{x}^{(t+1)}_i$};
\node[round_node]        (equal)       [right=0.5cm of target] {$=$};
\node[square_node_blue]      (term_1)       [right=0.5cm of equal] {$\mathbf{x}^{(t)}_i$};
\node[round_node]        (plus_1)       [right=0.5cm of term_1] {$+$};
\node[square_node_yellow]      (c_1)       [right=0.5cm of plus_1] {$\alpha$};
\node[square_node_blue]      (term_2)       [right=0.5cm of c_1] {$\left(\epsilon-\frac{1}{2}\right)$};
\node[round_node]        (plus_2)       [right=0.5cm of term_2] {$+$};
\node[square_node_yellow]      (c_2)       [right=0.5cm of plus_2] {$\beta$};
\node[square_node_blue]      (term_3)       [right=0.5cm of c_2] {$\left(\mathbf{g}^{(t)}-\mathbf{x}^{(t)}_i\right)$};

\node[box,label={\small \textbf{accPSO}},draw=blue!60,fit=(target)(term_3)] {};

\node[square_node_green,   below=1.2cm of target]           (target_5)  {$\mathbf{x}^{(t+1)}_i$};
\node[round_node]        (equal_5)       [right=0.5cm of target_5] {$=$};
\node[square_node_blue]      (term_51)       [right=0.52cm of equal_5] {$\mathbf{x}^{(t)}_i$};
\node[round_node]        (plus_51)       [right=0.48cm of term_51] {$+$};
\node[square_node_yellow]      (c_51)       [right=0.5cm of plus_51] {$\beta$};
\node[square_node_blue]      (term_52)       [right=0.5cm of c_51] {$\left(\mathbf{g}^{(t)}-\mathbf{x}^{(t)}_i\right)$};
\node[round_node]        (plus_52)       [right=0.5cm of term_52] {$+$};
\node[square_node_yellow]      (c_52)       [right=0.5cm of plus_52] {$\alpha$};
\node[square_node_blue]      (term_53)       [right=0.5cm of c_52] {$\left(\epsilon-\frac{1}{2}\right)$};

\node[box,label={\small \textbf{accFA}},draw=blue!60,fit=(target_5)(term_53)] {};

\node[square_node_green,   below=1.2cm of target_5]           (target_2)  {$\mathbf{x}^{(t+1)}_i$};
\node[round_node]        (equal_2)       [right=0.18cm of target_2] {$=$};
\node[square_node_yellow]      (w_21)       [right=0.18cm of equal_2] {$w$};
\node[square_node_blue]      (term_21)       [right=0.18cm of w_21] {$\mathbf{x}^{(t)}_i$};
\node[round_node]        (plus_21)       [right=0.18cm of term_21] {$+$};
\node[square_node_yellow]      (c_21)       [right=0.18cm of plus_21] {$c_1\cdot \epsilon_1$};
\node[square_node_blue]      (term_22)       [right=0.18cm of c_21] {$\left(\mathbf{g}^{(t)}-\mathbf{x}^{(t)}_i\right)$};
\node[round_node]        (plus_22)       [right=0.18cm of term_22] {$+$};
\node[square_node_yellow]      (c_22)       [right=0.18cm of plus_22] {$c_2\cdot \epsilon_2$};
\node[square_node_blue]      (term_23)       [right=0.18cm of c_22] {$\left(\mathbf{x}^{(*)}_i-\mathbf{x}^{(t)}_i\right)$};

\node[box,label={\small \textbf{PSO}},draw=blue!60,fit=(target_2)(term_23)] {};

\node[square_node_green,   below=1.2cm of target_2]           (target_3)  {$\mathbf{x}^{(t+1)}_i$};
\node[round_node]        (equal_3)       [right=0.5cm of target_3] {$=$};
\node[square_node_blue]      (term_31)       [right=0.58cm of equal_3] {$\mathbf{x}^{(t)}_i$};
\node[round_node]        (plus_31)       [right=0.42cm of term_31] {$+$};
\node[square_node_yellow]      (c_31)       [right=0.42cm of plus_31] {$\beta(r_i)$};
\node[square_node_blue]      (term_32)       [right=0.42cm of c_31] {$\left(\mathbf{x}^{(t)}_j-\mathbf{x}^{(t)}_i\right)$};
\node[round_node]        (plus_32)       [right=0.42cm of term_32] {$+$};
\node[square_node_yellow]      (c_32)       [right=0.42cm of plus_32] {$\alpha$};
\node[square_node_blue]      (term_33)       [right=0.42cm of c_32] {$\left(\epsilon-\frac{1}{2}\right)$};

\node[box,label={\small \textbf{FA}},draw=blue!60,fit=(target_3)(term_33)] {};

\node[square_node_green,   below=1.2cm of target_3]           (target_4)  {$\mathbf{x}^{(t+1)}_i$};
\node[round_node]        (equal_4)       [right=1.3cm of target_4] {$=$};
\node[square_node_blue]      (term_41)       [right=1.3cm of equal_4] {$\mathbf{x}^{(t)}_i$};
\node[round_node]        (plus_41)       [right=1.3cm of term_41] {$+$};
\node[square_node_yellow]      (q_4)       [right=1.32cm of plus_41] {$Q_i$};
\node[square_node_blue]      (term_42)       [right=1.28cm of q_4] {$\left(\mathbf{x}^{(t)}_i-\mathbf{g}^{(t)}\right)$};

\node[box,label={\small \textbf{BA (exploration)}},draw=blue!60,fit=(target_4)(term_42)] {};

\node[square_node_green,   below=1.2cm of target_4]           (target_6)  {$\mathbf{x}^{(t+1)}_i$};
\node[round_node]        (equal_6)       [right=1.32cm of target_6] {$=$};
\node[square_node_blue]      (term_61)       [right=1.32cm of equal_6] {$\mathbf{g}^{(t)}$};
\node[round_node]        (plus_61)       [right=1.32cm of term_61] {$+$};
\node[square_node_yellow]      (q_6)       [right=1.3cm of plus_61] {$\bar{r}$};
\node[square_node_blue]      (term_62)       [right=2.1cm of q_6] {$\mathcal{N}(0,1)$};

\node[box,label={\small \textbf{BA (exploitation)}},draw=blue!60,fit=(target_6)(term_62)] {};

\draw[->] (target.south) -- (target_5.north);
\draw[->] (target_5.south) -- (target_2.north);
\draw[->] (target_2.south) -- (target_3.north);
\draw[->] (target_3.south) -- (target_4.north);
\draw[->] (term_31.south) -- (term_41.north);
\draw[->] (term_1.south) -- (term_51.north);
\draw[->] (term_51.south) -- (term_21.north);
\draw[->] (term_53.south) -- (term_33.north);
\draw[->] (term_21.south) -- (term_31.north);
\draw[->] (term_22.south) -- (term_42.north);
\draw[->] (term_3.south) -- (term_52.north);
\draw[->] (term_2.south) -- (term_53.north);
\draw[->] (term_23.south) -- (term_32.north);
\draw[->] (term_52.south) -- (term_22.north);
\draw[->] (c_1.south) -- (c_52.north);
\draw[->] (c_2.south) -- (c_51.north);
\draw[->] (c_51.south) -- (c_21.north);
\draw[->] (c_52.south) -- (c_22.north);
\draw[->] (c_21.south) -- (c_31.north);
\draw[->] (c_22.south) -- (c_32.north);
\draw[->] (c_31.south) -- (q_4.north);
\draw[->] (q_4.south) -- (q_6.north);

\end{tikzpicture}
\caption{Simplified syntax diagrams of the variation operators used in the observed NI metaheuristic algorithms.}
\label{fig:architecture}
\end{figure}

\subsubsection{Variation operator analysis of the algorithms in the \textit{NIA}}
The variation operators analysis is dedicated to revealing how the variation operators as implemented by NI algorithm from \textit{NIA} are constructed, what common points they share, and how they diverge. The results of the comparative study are illustrated in Fig.~\ref{fig:architecture}, which highlights the structure of the variation operators in a form of simplified syntax diagrams, one for each algorithm.

From Fig.~\ref{fig:architecture} it can be observed, that although developing the observed NI algorithms are inspired by different NI metaphors, the implementations of the corresponding variation operators are founded on similar mathematical formulation. In general, each variation operator is expressed as a linear combination of a base factor and one/two additive terms. Furthermore, some similarities between the used additive terms conforming the variation operators can be exposed explicitly regarding the particular NI metaheuristics at hand. These similarities are denoted in the figure using directed arcs that connect two similar terms or factors. Those connections justify that the terms and factors emerging in the variation operators of different NI algorithms are equivalent mathematically.

The BNF notation is appropriate for describing the syntax of programming languages. In our study, the simplified BNF notation is employed for a description of the variation operators in the observed NI algorithms. The results of the variation operator analysis are presented in Tables~\ref{tab:bnf}-\ref{tab:bnf2}.

\begin{table}[htb]
    \caption{Simplified BNF notation of the variation operator syntax.}
    \label{tab:bnf}
    \centering
    \begin{tabular}{|l|}
    \hline
    $\langle \text{operator} \rangle ::= \langle \text{base\_term} \rangle~"+"~\langle\text{term\_1}\rangle~["+"~\langle \text{term\_2}] \rangle$ \\
    $\langle \text{base\_term} \rangle ::= [\langle\text{weight}\rangle]~[\langle\text{individual\_current}\rangle|\langle\text{global\_best}\rangle]$ \\
    $\langle \text{term\_1} \rangle ::= \langle\text{coefficient\_1}\rangle~[\langle\text{random\_term}\rangle|\langle\text{global\_best\_term}\rangle|\langle\text{FA\_neighbor\_term}\rangle|\langle\text{BA\_term}\rangle]$ \\
    $\langle\text{coefficient\_1}\rangle ::= [\langle \text{fixed}\rangle|\langle \text{stochastic}\rangle|\langle \text{deterministic}\rangle|\langle \text{metaphor-based}\rangle]$ \\
    $\langle\text{random\_term}\rangle::="("~\langle\text{epsilon}\rangle~"-"~\langle\text{one\_half}\rangle~")"$\\
    $\langle\text{global\_best\_term}\rangle::="("~\langle\text{global\_best}\rangle~"-"~\langle\text{individual\_current}\rangle~")"$\\
    $\langle\text{FA\_neighbor\_term}\rangle::="("~\langle\text{individual\_better}\rangle~"-"~\langle\text{individual\_current}\rangle")"$\\
    $\langle\text{BA\_term}\rangle::=\langle\text{global\_best}\rangle~"+"~\langle\text{metaphor\_based}\rangle\langle\text{epsilon\_normal}\rangle$\\
    $\langle\text{term\_2}\rangle::=\langle\text{coefficient\_2}\rangle~[\langle\text{global\_best\_term}\rangle|\langle\text{local\_best\_term}\rangle|\langle\text{random\_term}\rangle]$\\
    $\langle\text{coefficient\_2}\rangle ::= [\langle \text{fixed}\rangle|\langle \text{stochastic}\rangle]|\langle \text{deterministic}\rangle]$ \\
    $\langle\text{local\_best\_term}\rangle::="("~\langle\text{local\_best}\rangle~"-"~\langle\text{individual\_current}\rangle~")"$\\
    \hline
    \end{tabular}
\end{table}

\begin{table}[htb]
    \caption{Meaning of the undefined non-terminals.}
    \label{tab:bnf2}
    \centering
    \begin{tabular}{|ll|}
        \hline
        $\langle\text{weight}\rangle$ & $\text{Parameter}~w$ \\
        $\langle\text{fixed}\rangle$ & No parameter control \\
        $\langle\text{stochastic}\rangle$ & $\text{Parameter control involves randomness (e.g., } \epsilon_1\cdot c_1 \text{ or } \epsilon_2\cdot c_2 $) \\
        $\langle\text{deterministic}\rangle$ & Deterministic parameter control (e.g., $\alpha$ in FA) \\
        $\langle\text{metaphor-based}\rangle$ & Metaphor-based parameter control (e.g., $\beta(r)$ in FA, $\bar{r}$ in BA) \\
        $\langle\text{epsilon}\rangle$ & $\text{Random number drawn from uniform distribution}~\epsilon\in\mathcal{U}(0,1)$ \\
        $\langle\text{mean\_pulse\_rate}\rangle$ & Average pulse rate \\
        $\langle\text{epsilon\_normal}\rangle$ & Random number drawn from Gaussian distribution $\mathcal{N}(0,1)$\\
        $\langle\text{one\_half}\rangle$ & $\text{Fraction}~\frac{1}{2}$ \\
        $\langle\text{global\_best}\rangle$ & $\text{Global best solution}~\textbf{g}^{(t)}$ \\
        $\langle\text{local\_best}\rangle$ & $i-\text{th local best solution}~\textbf{x}^*_i$ \\
        $\langle\text{individual\_current}\rangle$ & $\text{Current individual}~\textbf{x}^{(t)}_i$ \\
        $\langle\text{individuals\_better}\rangle$ & $\text{New vector}~\forall \mathbf{x}^{(t)}_j\in\mathcal{N}(\textbf{x}^{(t)}_i):\textbf{x}^{(t+1)}_i=\textbf{x}^{(t)}_i+\beta(\textbf{x}^{(t)}_j-\textbf{x}^{(t)}_i)$,  \\
        & $~\text{where}~\mathcal{N}(\textbf{x}^{(t)}_i)=\{\forall\mathbf{x}^{(t)}_j:f(\mathbf{x}^{(t)}_j)<\mathbf{x}^{(t)}_i\}$\\
        \hline
    \end{tabular}
\end{table}

The main advantage of using the simplified BNF notation is the simplicity of describing the various variation operators in a universal notation. This fact justifies that all the variation operators in NI algorithms need to have the same origins.

\subsection{Experimental setup}
As a test bed, a Sphere function~\cite{Naser2025review} was selected that is defined as:
\begin{equation}
    f(\textbf{x})=-\sum_{i=1}^D x^2_i,
\end{equation}
where $D$ designates the function dimension. 

In our experimental study, the dimension of function was set to $D=10$, while all the algorithms shared some parameters with the same meaning. These parameters were set to identical values to make the comparison between the NI algorithms as fair as possible. The shared parameters, remaining fixed during the runs, were set as follows: the same population size $\mathit{Np}=20$, as well as the same termination condition, i.e., the maximum number of generations $T=500$. Although the problem dimension is not large in the sense of the NI algorithms, the strong equivalence is set rigorously enough for identifying the behavior of two algorithms already with smaller dimensions. Even with these conservative values, the size of the feature vector is huge, even under these limited conditions, precisely $|\mathit{FV}|=4\cdot (500+20)=2080$. Consequently, this required $500\times20=10,000$ independent runs of the controlled algorithm using different hyper-parameter settings for each initialization seed.
    
The experiments using the proposed framework were conducted as follows: At first, the control algorithm was selected for optimizing the Sphere function. All the control algorithms were executed $N=151$ times using different seeds. Next, the controlled algorithms, already integrated into the framework, were selected sequentially one at a time from the \textit{NIA} set for different seeds. Thereby, the \texttt{Meta-DE} varied the hyper-parameters within the domains displayed in Table~\ref{tab:param_set}. In summary, 151 different feature vectors were obtained, that demand $151\times 10,000=1,51$~M independent runs launched by the \texttt{Meta-DE} control system per single controlled algorithm. The characteristics of the computer system on which the experiments were executed are presented in Table~\ref{tab:HW}. 

\begin{table*}[htb]
    \small
    \caption{Characteristics of the PC and the correlation coefficient interpretation.}
    \label{tab:main_tab}
    \begin{subtable}[h]{0.44\textwidth}
    \centering
    \begin{tabular}{l|l}
    \hline
    Computer component & Type \\    
    \hline\hline
    Operating system & Linux Mint 21.3  \\
    Processor &  AMD Ryzen 7 5700G \\
    Memory & 15.2 GB \\
    Graphic processor unit & NVIDIA GeForce RTX \\
    Hard disk drive & 10 TB \\
    \hline
    \end{tabular}
    \caption{Characteristics of the personal computer.}
    \label{tab:HW}
    \end{subtable}
    \hfill
    \begin{subtable}[h]{0.42\textwidth}
    \centering
    \begin{tabular}{l|l}
    \hline
    Value & Corr. interpretation \\
    \hline\hline
    $[0.00‑0.19]$ & very weak\\
    $[0.20‑0.39]$ & weak\\
    $[0.40‑0.59]$ & moderate\\
    $[0.60‑0.79]$ & strong\\
    $[0.80‑1.00]$ & very strong\\
    \hline
    \end{tabular}
    \caption{Interpretation of the corresponding coefficients.}
    \label{tab:corr_coef}
     \end{subtable}
\end{table*}

After determining the feature vectors of the controlled algorithms with the highest operating resemblance to the control algorithm, we performed a statistical analysis, in which were applied statistical and ML-based similarity metrics. The values of the calculated statistical metrics obtained by the statistical metrics $Sim_{\cos}$, $Sim_{\text{SMAPE}}$, and $\rho$ are interpreted in the study as illustrated in Table~\ref{tab:corr_coef}.

We also used three ML-based classifiers to test for separability of the algorithms' behaviors, namely the k-Nearest Neighbors (kNN)~\cite{Fix1989dicriminatory}, Support Vector Machines (SVM)~\cite{Cortes1995support}, and a Random Forest (RF)~\cite{Ho1995random}. 

Specifically, the reference and the controlled algorithm feature vectors obtained with the same seed (i.e., the initial population) were paired, resulting in $N=151$ pairs of feature vectors representing the training set for the ML classifiers. We used a kNN classifier with a neighborhood size of $k=5$, an SVM with a radial basis kernel and regularization weight $C=1$, and a random forest with $100$ trees and a maximum tree depth of $5$. A 10-fold cross-validation was then employed to evaluate the classifiers' efficiency in discerning the algorithms, where the average classification accuracy $A$ was used as an indicator of algorithm (dis)similarity (i.e., the algorithm similarity was defined as $1-A$).

The following four case studies were performed in order to approve/deny the research questions posted at the beginning of the section:
\begin{itemize}
    \item the \texttt{accPSO} and \texttt{accFA} are strongly equivalent,
    \item the metaphors affect the implementation of the NI metaheuristics crucially,
    \item creating conditions for achieving equivalence between two NI metaheuristics are hard,
    \item the \texttt{BA} is \texttt{PSO}.
\end{itemize}
In the remainder of this section, the findings of the mentioned case studies are reported in detail. The section concludes with a discussion of the obtained results. 

\subsection{Case Study 1: The \texttt{accPSO} and \texttt{accFA} are strongly equivalent}
The goal of Case Study~1 was to approve \textbf{Hypothesis~\ref{hypo:1}}. In line with this, the necessary tasks for drawing conclusions by means of the proposed framework are described in the remainder of the paper.

\subsubsection{CS-1: Selection of algorithm's test set}
The set of observed algorithms was defined as $\mathit{NIA}^{(\text{CS-1})}=\{\texttt{accPSO,accFA}\}$. As a control algorithm, the \texttt{accPSO} was selected, while the reference feature vectors were obtained using the parameters $\alpha=0.5$ and $\beta=0.2$. 

\subsubsection{CS-1: Results of the parameter setting analysis}
The implementations of both accelerated algorithms (i.e., \texttt{accPSO} and \texttt{accFA}) were the same according to the mathematical theory, as is evident by the components' analysis. Therefore, the \texttt{Meta-DE} focuses on searching for those parameter settings of the controlled algorithms a\texttt{ccPSO} and \texttt{accFA} that are the most equivalent to the control algorithm \texttt{accPSO} parameter settings (i.e., $\alpha=0.5$ and $\beta=0.2$). The results of the Meta-DE are shown in Table~\ref{tab:cs1}, that is divided into 'Control' and 'Controlled' parts: The former were obtained by the \texttt{Meta-DE} control system by tuning the \texttt{accPSO} algorithm parameters initialized randomly, while the latter by tuning the \texttt{accFA} algorithm parameters. The parameters are enumerated in the column 'Metrics'. Besides the results of parameter tuning, the fitness function values are included into the table for each of the observed algorithms, and the quality of the observed optimization algorithm is highlighted. All the mentioned quality metrics were estimated using statistical measures, like: minimum, maximum, mean, standard deviation, and median obtained after 151 independent runs.
\begin{table}[htb]
    \caption{The results of the \texttt{Meta-DE} by parameter setting analysis in CS-1.}
    \label{tab:cs1}
    \centering
    \begin{tabularx}{\textwidth}{l|l|l|
        >{\raggedleft}p{1.3cm}
        >{\raggedleft}p{1.3cm}
        >{\raggedleft}p{1.3cm}
        >{\raggedleft}p{1.3cm}
        >{\raggedleft}p{1.3cm}}
    \hline
    Role & Algorithm & Metric & Min & Max & Mean & StDev & Median\\
    \toprule
    \multirow{3}{*}{\resizebox{.16cm}{!}{\rotatebox{90}{Control}}} & \multirow{3}{*}{\texttt{accPSO}} & $\alpha$ & 0.4999 & 0.5001 & 0.5000 & $<0.0001$ & 0.5000 \\
     & & $\beta$ & 0.2000 & 0.2000 & 0.2000 & 0.0000 & 0.2000 \\
     \cmidrule{3-8}
     & & Fitness & -0.1197 & -0.0134 & -0.0618 & 0.0269 & -0.0577 \\
     \cmidrule{1-8}
    \multirow{3}{*}{\resizebox{.16cm}{!}{\rotatebox{90}{Controlled}}} & \multirow{4}{*}{\texttt{accFA}} & $\alpha$ & 0.4999 & 0.5001 & 0.5000 & $<0.0001$ & 0.5000 \\
     & & $\beta$ & 0.2000 & 0.2000 & 0.2000 & 0.0000 & 0.2000 \\
     \cmidrule{3-8}
     & & Fitness & -0.1197 & -0.0134 & -0.0618 & 0.0269 & -0.0577 \\
    \hline     
    \end{tabularx}
\end{table}

In summary, the parameter setting analysis reveals that both accelerated algorithms applied the same number of parameters. Although the parameter $\alpha$ and $\beta$ preceding two additive terms in the variation operator equation arising in the opposite order, the parameter settings converged to the same values in both accelerated algorithms after tuning. Also the behavior of the \texttt{accPSO} in the phenotype space is strongly equivalent to the \texttt{accFA} as can be seen from the similar values of metrics in the table.

\subsubsection{CS-1: Results of the statistical analysis for identifying the strong equivalence of NI algorithms}

The strong equivalence of NI algorithms was identified in a statistical analysis, where the pair of control/controlled feature vectors were compared according to the similarity metrics. In this study, the \texttt{accPSO} emerged as the control algorithm, and the \texttt{accFA} as controlled ones. The results are presented in Table~\ref{tab:cs1-detail}, that illustrates the values of the various similarity metrics for each controlled algorithm. In our case, only the results of the \texttt{accFA} controlled algorithm are presented in the table. Due to their stochastic nature, each parameter is exhibited by its corresponding statistical measures, like: minimum (tag 'Min' in the table), maximum (tag 'Max' in the table), mean (tag 'Mean' in the table), standard deviation (tag 'StDev' in the table), and median (tag 'Median' in the table).
\begin{table}[htb]
    \caption{The results of the statistical analysis according to similarity metrics for identifying the strong equivalence between the \texttt{accPSO} and \texttt{accFA} in CS-1.}
    \label{tab:cs1-detail}
    \centering
    \begin{tabularx}{\textwidth}{l|l|
        >{\raggedleft}p{1.1cm}
        >{\raggedleft}p{1.1cm}
        >{\raggedleft}p{1.1cm}
        >{\raggedleft}p{1.1cm}
        >{\raggedleft}p{1.1cm}
        >{\raggedleft}p{1.1cm}}
    \hline
    Algorithm & Parameter & $Sim_{\cos}$ & $Sim_{\text{SMAPE}}$ & $\rho$ & $Sim_{\text{kNN}}$ & $Sim_{\text{SVM}}$ & $Sim_{\text{RF}}$ \\
    \toprule
    \multirow{5}{*}{accFA} & Min & 1.0000 & 1.0000 & 1.0000 & n/a & n/a & n/a \\
     & Max & 1.0000 & 1.0000 & 1.0000 & n/a & n/a & n/a \\
     & Mean & \textbf{1.0000} & \textbf{1.0000} & \textbf{1.0000} & \textbf{1.0000} & \textbf{1.0000} & \textbf{1.0000} \\
     & StDev & \textbf{0.0000} & \textbf{0.0000} & \textbf{0.0000} & \textbf{0.0000} & \textbf{0.0000} & \textbf{0.0000} \\
     & Median & 1.0000 & 1.0000 & 1.0000 & n/a & n/a & n/a \\
    \hline     
    \end{tabularx}
\end{table}

As is evident from the table, all the values of the statistical similarity metrics (i.e., $Sim_{\cos}$,$Sim_{\text{SMAPE}}$, and $\rho$) equate to one, indicating the very strong correlation of the corresponding feature vectors, while the corresponding standard deviation to zero. On the other hand, the classification accuracies by the ML methods used (i.e., $Sim_{\text{kNN}}$, $Sim_{\text{SVM}}$, and  $Sim_{\text{RF}}$) are set to $1.0\pm0.0$ indicating that the ML methods cannot classify patterns (i.e., feature vectors) with certainty. This justified that the \texttt{accPSO} and \texttt{accFA} are strongly equivalent.

\subsubsection{CS-1: Accepting/rejecting the hypothesis}
\textbf{Hypothesis~1} can be approved due to the very strong correlation achieved by the statical similarity metrics, as well as the classification indeterminacy indicated by the ML classification methods. Interestingly, the framework revealed an additional dilemma: ''The \texttt{accPSO} and \texttt{accFA} used different metaphors in the development, but their implementations gave the strong equivalence behavior''.

\subsection{Case Study 2: The metaphors provided the inspiration for the implementation of several metaheuristics components}
The goal of this case study was to show that the metaphors do not only affect the implementation of the variation operators (i.e., the \textsc{Move} function in SI-based algorithms), but also the other algorithm components, by developing the NI metaheuristics and how big these effects are. In accordance with this, \textbf{Hypothesis~\ref{hypo:2}} was posted. Using the framework, this hypothesis testing demands background researches as illustrated in Section~\ref{sec:deep}. In the remainder of the paper, the testing tasks are described more detail.

\subsubsection{CS-2: Selection of algorithm's test set}
In this case study, the set of observed NI algorithms was extended and defined as $\mathit{NIA}^{(\text{CS-2})}=\{\texttt{accPSO,PSO,FA,BA}\}$. In accordance with the conclusions of Case Study~1, where we found that both accelerated algorithms are strongly equivalent, only the \texttt{accPSO} was taken into consideration here. As a control algorithm, the mentioned accelerated algorithm was selected, while the reference feature vectors were obtained using the same parameter settings as in the last case study, i.e., $\alpha=0.5$ and $\beta=0.2$. 

\subsubsection{CS-2: Results of the parameter setting analysis}
The results of the parameter setting analysis are depicted in Table~\ref{tab:cs2}, where the parameter settings tuned by the \texttt{Meta-DE} and the corresponding fitness function values obtained by the control \texttt{accPSO} are compared by the same ones produced by the particular controlled algorithm from the set $\mathit{NIA}^{(\text{CS-2})}\backslash\mathtt{accPSO}$ according to the statistical measures.
\begin{table}[htb]
    \caption{The results of the \texttt{Meta-DE} by parameter setting analysis in CS-2.}
    \label{tab:cs2}
    \centering
{\footnotesize
    \begin{tabularx}{\textwidth}{l|l|l|rrrrr}
    \hline
    Role & Algorith. & Metric & Min & Max & Mean & StDev & Median\\
    \midrule
    \multirow{3}{*}{\rotatebox{90}{Control}} & \multirow{3}{*}{\texttt{accPSO}} & $\alpha$ & \textbf{0.4999} & \textbf{0.5001} & 0.5000 & $\mathbf{<0.0001}$ & 0.5000\\
     & & $\beta$ & 0.2000 & 0.2000 & 0.0000 & 0.2000 & 0.2000 \\
     \cmidrule{3-8}
     & & Fitness & \textbf{-0.1197} & \textbf{-0.0134} & -0.0618 & \textbf{0.0269} & -0.0577 \\
     \cmidrule{1-8}
    \multirow{13}{*}{\rotatebox{90}{Controlled}} & \multirow{4}{*}{\texttt{PSO}} & $w$ &  0.4740 & 0.7760 & 0.6322 & 0.0780 & 0.6251 \\ 
     & & $c_1$ & 1.5000 & 1.8624 & 1.5511 & 0.1084 & 1.5000 \\
     & & $c_2$ & \textbf{1.5000} & \textbf{2.5000} & 2.3575 & \textbf{0.2193} & 2.4712 \\
     \cmidrule{3-8}
     & & Fitness & -1.8743 & $<0.0001$ & -0.0820 & 0.3739 & $<0.0001$ \\ 
     \cmidrule{2-8}
     & \multirow{4}{*}{\texttt{FA}} & $\alpha$ & 0.3825 & 1.0000 & 0.9322 & 0.9503 & 0.0744 \\
     & & $\beta$ & \textbf{0.5000} & \textbf{1.5000} & 1.0914 & \textbf{1.1369} & 0.3533 \\
     & & $\gamma$ & \textbf{0.1000} & \textbf{1.0000} & 0.3953 & \textbf{0.2297} & 0.3302 \\
     \cmidrule{3-8}
     & & Fitness & \textbf{-10012.2000} & \textbf{-2135.9100} & -6267.5660 & \textbf{-6194.1100} & 1344.8668 \\
     \cmidrule{2-8}
     & \multirow{3}{*}{\texttt{BA}} & $A$ & 1.1311 & 1.2000 & 1.1570 & 0.0158 & 1.1497 \\
     & & $\gamma$ & \textbf{0.1000} & \textbf{0.3000} & 0.2342 & \textbf{0.0695} & 0.2453 \\
     \cmidrule{3-8}
     & & Fitness & \textbf{-666.5810} & \textbf{-111.2480} & -275.0908 & \textbf{122.3454} & -247.3120 \\
     \hline     
     \end{tabularx}
}     
\end{table}

As can be seen in Table~\ref{tab:cs2}, the parameter $\alpha$ of the control algorithm \texttt{accPSO}, tuned by the \texttt{Meta-DE}, showed some variability during various independent runs, although this variability was negligible. On the other hand, the \texttt{Meta-DE} found the optimal values for some parameters of the controlled algorithms that are located at the lower and upper bounds of their domains in some independent runs, e.g.: the \texttt{PSO} parameter $c_2$, the \texttt{FA} parameters $\beta$ and $\gamma$, and the \texttt{BA} parameter $r$. These parameters also exhibit a higher variance. On the other hand, the fitness values showed that the observed algorithms are of different qualities, because they are very dissimilar, when they are compared between each other regarding to the statistical measures, like minimum, maximum and standard deviation. The results could be confusing, however, the goal of the framework is not to find the best fitness function values, but the most similar behavior of the NI algorithms.

\subsubsection{CS-2: Results of identifying the strong equivalence of the NI algorithms}
The results of the statistical analysis for identifying the strong identity of the control algorithm \texttt{accPSO} and controlled algorithms drawn from the set $\mathit{NIA}^{(\text{CS-2})}\backslash\texttt{accPSO}$ are depicted in Table~\ref{tab:cs2-detail}. 
\begin{table}[htb]
    \caption{The results of the statistical analysis according to the similarity metrics for identifying the strong equivalence using the \texttt{accPSO} as the control algorithm in CS-2.}
    \label{tab:cs2-detail}
    \centering
    \begin{tabularx}{\textwidth}{l|l|rrrrrr}
    \hline
    Algorithm & Parameter & $Sim_{\cos}$ & $Sim_{\text{SMAPE}}$ & $\rho$ & $Sim_{\text{kNN}}$ & $Sim_{\text{SVM}}$ & $Sim_{\text{RF}}$ \\
    \midrule
    \multirow{5}{*}{\texttt{PSO}} & Min & 0.9674 & 0.9997 & 0.9673 & n/a & n/a & n/a \\
     & Max & \textbf{0.9818} & \textbf{0.9999} & \textbf{0.9819} & n/a & n/a & n/a \\
     & Mean & \textbf{0.9756} & \textbf{0.9998} & \textbf{0.9756} & \textbf{0.0000} & \textbf{0.0000} & \textbf{0.0000} \\
     & StDev & \textbf{0.0038} & $\mathbf{<0.0001}$ & \textbf{0.0038} & \textbf{0.0000} & \textbf{0.0000} & \textbf{0.0000} \\
     & Median & 0.9764 & 0.9998 & 0.9764 & n/a & n/a & n/a \\
    \hline
    \multirow{5}{*}{\texttt{FA}} & Min & 0.1410 & 0.9991 & 0.1351 & n/a & n/a & n/a \\
     & Max & 0.3954 & 0.9991 & 0.3752 & n/a & n/a & n/a \\ 
     & Mean & \textbf{0.2058} & \textbf{0.9991} & \textbf{0.1986} & \textbf{0.0000} & \textbf{0.0000} & \textbf{0.0000} \\
     & StDev & \textbf{0.0460} & \textbf{0.0000} & \textbf{0.0430} & \textbf{0.0000} & \textbf{0.0000} & \textbf{0.0000} \\
     & Median & 0.1968 & 0.9991 & 0.1904 & n/a & n/a & n/a \\
     \hline
    \multirow{5}{*}{\texttt{BA}} & Min & 0.7273 & 0.9993 & 0.7315 & n/a & n/a & n/a \\
     & Max & 0.7391 & 0.9994 & 0.9994 & n/a & n/a & n/a \\
     & Mean & \textbf{0.8231} & \textbf{0.9994} & \textbf{0.8259} & \textbf{0.0070} & \textbf{0.0030} & \textbf{0.0000} \\
     & StDev & \textbf{0.0546} & \textbf{0.0001} & \textbf{0.0539} & \textbf{0.0130} & \textbf{0.0100} & \textbf{0.0000} \\
     & Median & 0.8254 & 0.9994 & 0.8311 & n/a & n/a & n/a \\
     \hline
     \end{tabularx}
\end{table}

As can be seen in the table, the original \texttt{PSO} algorithm achieved the best results according to the statistical metrics. This means that, in the best case, these metrics gained values indicating a ''very strong'' correlation with the \texttt{accPSO} algorithm. The same conclusion is also valid in the average case. The strong correlation is confirmed by low variability (i.e., a lower standard deviation). 

On the other hand, worse results according to the statistical similarity metrics were detected by the \texttt{FA} controlled algorithm, where the reported values indicate ''low'' and ''very low'' correlation with the control algorithm \texttt{accPSO}. The situation is better when the results of the \texttt{BA} controlled algorithm are shown, because its reported values of statistical similarity metrics indicates ''strong'' correlation with the \texttt{accPSO}. The results according to the similarity statistical metric $Sim_{\text{SMAPE}}$ showed a trend that the absolute percentage error converges to zero for all the observed algorithms. This trend has origins in the definition of the similarity metric, and it is a consequence of optimization due to the \texttt{Meta-DE}. 

According to the ML similarity metrics, the strong equivalence with the control \texttt{accPSO} cannot be confirmed by neither controlled algorithm in the test. 

\subsubsection{CS-2: Accepting/rejecting the hypothesis}

\textbf{Hypothesis~\ref{hypo:2}} asserts that metaphors provide an inspiration for implementation of several metaheuristic components. Unfortunately, the component analysis of the algorithms in $\mathit
{NIA}^{(\text{CS-2})}$ in Table~\ref{tab:components} revealed the opposite: Indeed, the main difference between the observed NI algorithms is indicated in the implementation of the variation operators (i.e., the formula governing the \textsc{Move} operator). The algorithms \texttt{PSO} and \texttt{BA} present exceptions in the table. The former introduces the local best population of individuals, on the one hand, which demands an additional replacement selection scheme 'one-to-one', on the other. The latter introduces the explicit balancing between exploration/exploitation controlled by the parameter pulse rate $r_i$.

Primarily, the metaphors influence the implementation of the variation operators (Table~\ref{tab:bnf3}).
\begin{table}[htb]
    \caption{The comparison of the variation operators from the set $\mathbf{NIA}^{(\text{CS-2})}$.}
    \label{tab:bnf3}
    \centering
    {\footnotesize
    \begin{tabular}{|l|}
    \hline
    $\langle \text{oper\_accPSO} \rangle ::= \langle \text{base\_term} \rangle~"+"~\langle\text{fixed}\rangle\langle \text{random\_term} \rangle~"+"~\langle\text{fixed}\rangle\langle \text{global\_best\_term} \rangle$ \\
    \hline\hline
    $\langle \text{oper\_accFA} \rangle ::= \langle \text{base\_term} \rangle~"+"~\langle\text{fixed}\rangle\langle \text{global\_best\_term} \rangle~"+"~\langle\text{fixed}\rangle\langle \text{random\_term} \rangle$ \\
    $\langle \text{oper\_PSO} \rangle ::= \langle w\rangle\langle \text{base\_term} \rangle~"+"~\langle\text{stochastic}\rangle\langle \text{global\_best\_term} \rangle~"+"~\langle\text{stochastic}\rangle\langle \text{local\_best\_term} \rangle$ \\
    $\langle \text{oper\_FA} \rangle ::= \langle \text{base\_term} \rangle"+"\langle\text{metaphor\_based}\rangle\langle \text{FA\_neighbor\_term} \rangle"+"\langle\text{deterministic}\rangle\langle \text{random\_term} \rangle$ \\
    $\langle \text{oper\_BA\_exploration} \rangle ::= \langle \text{base\_term} \rangle~"+"~\langle\text{stochastic}\rangle\langle \text{global\_best\_term}\rangle$ \\
    $\langle \text{oper\_BA\_exploitation} \rangle ::= \langle \text{base\_term} \rangle~"+"~\langle\text{metaphor\_based}\rangle\langle \text{epsilon\_normal}\rangle$ \\
    \hline
    \end{tabular}
    }
\end{table}

As can be seen from the table, all these implementation of various variation operators share a similar structure, which can be described using a simple BNF notation. This means that, actually, the metaphors are only seemingly related to the implementation of the NI metaheuristics. Moreover, the non-terminal $\langle\text{global\_best\_term}\rangle$ arose in four from the five variation operators of the observed NI metaheuristics. This means that various metaphors are implemented in a similar manner.

Consequently, \textbf{Hypothesis~\ref{hypo:2}} must be rejected.

\subsection{Case Study 3: Creating conditions for achieving a strong equivalence between two NI metaheuristics is hard}
The purpose of \textbf{Case study~3} was to approve/deny \textbf{Hypothesis-\ref{hypo:3}}, asserting that computing conditions in which the strong equivalence of NI algorithms can be proven, are not easy to find. To confirm the hypothesis, the hybrid version of \texttt{FA} (the so-called \texttt{FAv2}) was implemented by the authors of the paper, in order to show that improving the results of the original \texttt{FA} can increase the equivalence between the other NI algorithms, and the \texttt{FAv2} is equipped with a 'one-to-one' selection mechanism. As illustrated in Algorithm~\ref{alg:fa2}, 
\begin{algorithm}[htb]
\caption{Improved Firefly algorithm \texttt{FAv2}}
\label{alg:fa2}
\begin{algorithmic}[1]
\Require randomization $\alpha$, attractiveness $\beta_0$, absorption coefficient $\gamma$
\State $t = 0; \mathbf{x}_{\mathit{best}} = \emptyset;$ \Comment {initialization of parameters}
\State $P^{(0)}$ = \textsc{Initialize}; \Comment {initialization of population}
\State $\textit{MP}^{(0)}$ = $P^{(0)}$; \Comment {initialization of the mating pool}
\While {$\mathbf{not}$ \textsc{TerminationConditionMeet}}
\ForAll{$\mathbf{y}_i \in \textit{MP}^{(t)}$}
\State $f_{trial}$ = \textsc{Evaluate}($\mathbf{y}^{(t)}$); \Comment {evaluate $\mathbf{y}_i\in \textit{MP}^{(t)}$ regarding the $f(\mathbf{y}_i)$}
\If {$f_{trial} > f_i)$} \Comment {replace with the better solution} 
\State $\mathbf{x}_i = \mathbf{y}_i;f_i=f_{trial};$
\If {$f_{best} > f_i$} \Comment {determine the best solution} 
\State $\mathbf{x}_{best} = \mathbf{x}_i;f_{best}=f_i;$ 
\EndIf
\EndIf
\EndFor
\ForAll{$\mathbf{x}_i \in P^{(t)}$}
\ForAll{$\mathbf{x}_j \in P^{(t)}$}
\If {$f(\mathbf{x}_j) > f(\mathbf{x}_i)$}
\State $r$=\textsc{Euclidean}$(\mathbf{x}_i,\mathbf{x}_j)$; \Comment calculate Euclidean distance 
\State $\mathbf{y}_i$=\textsc{Move}$(\mathbf{x}_i,\mathbf{x}_j,\alpha,\beta_0*\exp^{-\gamma r^2})$; \Comment {move firefly $\mathbf{x}_i$ according Eq.~(\ref{eq:fa})}
\EndIf
\EndFor
\EndFor
\State $\alpha = \alpha*0.98$; \Comment{decay factor - tweak as needed}
\State $t = t+1$;
\EndWhile
\end{algorithmic}
\end{algorithm}

the selection mechanism increases the selection pressure of the original \texttt{FA}, and introduces a 2-population model (elitist population and mating pool) by the \texttt{FAv2}. 

\subsubsection{CS-3: Selection of the algorithm's test set}
The set of the NI metaheuristic algorithms in the case study was defined as $\mathit{NIA}^{(\text{CS-3})}=\{\texttt{accPSO,PSO,FA,FAv2,BA}\}$. The original \texttt{PSO} algorithm was taken as a control algorithm using the following parameter settings: $w=0.5$, $c1=1.5$, and $c2=1.5$ by generating the reference feature vectors. 

\subsubsection{CS-3: Results of the parameter setting analysis}
The results of the parameter setting analysis are depicted in Table~\ref{tab:cs3}, where the parameter setting tuned by the \texttt{Meta-DE} and the corresponding fitness function value obtained by the control \texttt{PSO} are compared by the same ones produced by the particular controlled algorithm from the set $\mathit{NIA}^{(\text{CS-3})}\backslash\mathtt{accPSO}$ according to the statistical measures.
\begin{small}
\begin{table}[htb]
    \caption{The results of the \texttt{Meta-DE} by parameter setting analysis in CS-3.}
    \label{tab:cs3}
    \centering
    \begin{tabularx}{\textwidth}{l|l|l|rrrrr}
    \hline
    Role & Algorit. & Metric & Min & Max & Mean & StDev & Median\\
    \midrule
    \multirow{4}{*}{\rotatebox{90}{Control}} & \multirow{3}{*}{\texttt{PSO}} & $w$ & \textbf{0.4954} & \textbf{0.5000} & 0.4999 & \textbf{0.0004} & 0.5000 \\
     & & $c_1$ & \textbf{1.5000} & \textbf{1.5002} & 1.5000 & $\mathbf{<0.0001}$ & 1.5000 \\
     & & $c_2$ & \textbf{1.5000} & \textbf{1.5106} & 1.5001 & \textbf{0.0009} & 1.5000 \\
     \cmidrule{3-8}
     & & Fitness & $<-0.0001$ & $<-0.0001$ & $\mathbf{<-0.0001}$ & $<0.0001$ & $<-0.0001$ \\
     \cmidrule{1-8}
    \multirow{18}{*}{\rotatebox{90}{Controlled}} & \multirow{4}{*}{\texttt{accPSO}} & $\alpha$ & 0.1696 & 1.0000 & 0.9027 & 0.2185 & 1.0000 \\
     & & $\beta$ & 0.1301 & 0.1936 & 0.1513 & 0.0109 & 0.1502 \\
     \cmidrule{3-8}
     & & Fitness & -0.5043 & -0.0088 & \textbf{-0.2686} & 0.1174 & -0.2851 \\
     \cmidrule{2-8}
     & \multirow{4}{*}{\texttt{FA}} & $\alpha$ & 0.6919 & 1.0000 & 0.9432 & 0.0569 & 0.9611 \\
     & & $\beta$ & \textbf{0.5000} & \textbf{1.5000} & 1.1502 & \textbf{0.3486} & 1.2396 \\
     & & $\gamma$ & \textbf{0.1000} & \textbf{1.0000} & 0.3940 & \textbf{0.2974} & 0.3249 \\
     \cmidrule{3-8}
     & & Fitness & -9508.9300 & -2303.9400 & \textbf{-6420.3970} & 1279.0872 & -6453.3500 \\
     \cmidrule{2-8}
     & \multirow{4}{*}{\texttt{FAv2}} & $\alpha$ & 0.3109 & 0.6364 & 0.4412 & 0.0703 & 0.4310 \\
     & & $\beta$ & \textbf{0.5000} & \textbf{1.5000} & 1.2034 & \textbf{0.2711} & 1.2597 \\
     & & $\gamma$ & \textbf{0.1000} & \textbf{1.0000} & 0.1322 & \textbf{0.1136} & 0.1000 \\
     \cmidrule{3-8}
     & & Fitness & -0.0553 & $<0.0001$ & \textbf{-0.0009} & 0.0063 & $<0.0001$ \\
     \cmidrule{2-8}
     & \multirow{3}{*}{\texttt{BA}} & $A$ & 1.0940 & 1.2000 & 1.1475 & 0.0177 & 1.1474 \\
     & & $\gamma$ & \textbf{0.1000} & \textbf{0.3000} & 0.2409 & \textbf{0.0570} & 0.2549 \\
     \cmidrule{3-8}
     & & Fitness & -611.1540 & -35.6280 & \textbf{-224.0569} & 108.8632 & -205.9640 \\
     \hline     
    \end{tabularx}
\end{table}
\end{small}   
As is evident from the table, the best parameter setting of the control algorithm as found by the \texttt{Meta-DE} exhibited some variances. This fact shows that these parameters are sensitive to smaller variations in their values, but do not influence the quality of the results crucially. On the one hand, the \texttt{Meta-DE} searched for the best values of some parameters in the whole interval of the domain during all the independent runs. These values are presented emboldened in the table. On the other hand, all the observed algorithms achieved different results in the phenotype space, where the \texttt{PSO} produced the best ones that are closer to zero ($<-0.0001$). The other algorithms achieved results far from the optimum. Indeed, the hybrid version \texttt{FAv2} improved the results of their original counterpart FA as well as \texttt{accPSO} and \texttt{BA} substantially.

\subsubsection{CS-3: Results of identifying the strong equivalence of NI algorithms}
The results of the statistical analysis for identifying the strong equivalence of the control algorithm \texttt{PSO} and controlled algorithms drawn from the set $\mathit{NIA}^{(\text{CS-3})}\backslash\texttt{PSO}$ are depicted in Table~\ref{tab:cs3-detail}. 
\begin{table}[htb]
    \caption{The results of the statistical analysis according to the similarity metrics for identifying the strong equivalence between the \texttt{PSO} as control algorithm in CS-3.}
    \label{tab:cs3-detail}
    \centering
    \begin{tabularx}{\textwidth}{l|l|rrrrrr}
    \hline
    Algorithm & Parameter & $Sim_{\cos}$ & $Sim_{\text{SMAPE}}$ & $\rho$ & $Sim_{\text{kNN}}$ & $Sim_{\text{SVM}}$ & $Sim_{\text{RF}}$ \\
    \midrule
    \multirow{5}{*}{\texttt{accPSO}} & Min & 0.9549 & 0.9997 & 0.9561 & n/a & n/a & n/a \\
     & Max & \textbf{0.9924} & \textbf{0.9999} & \textbf{0.9924} & n/a & n/a & n/a \\
     & Mean & \textbf{0.9862} & \textbf{0.9999} & \textbf{0.9863} & \textbf{0.0000} & \textbf{0.0000} & \textbf{0.0000} \\
     & StDev & \textbf{0.0045} & \textbf{0.0000} & \textbf{0.0044} & \textbf{0.0000} & \textbf{0.0000} & \textbf{0.0000} \\
     & Median & 0.9869 & 0.9999 & 0.9869 & n/a & n/a & n/a \\
    \hline
    \multirow{5}{*}{\texttt{FA}} & Min & 0.1541 & 0.9991 & 0.1499 & n/a & n/a & n/a \\
     & Max & 0.2080 & 0.9991 & 0.2036 & n/a & n/a & n/a \\
     & Mean & \textbf{0.1827} & \textbf{0.9991} & \textbf{0.1789} & \textbf{0.0000} & \textbf{0.0000} & \textbf{0.0000} \\
     & StDev & \textbf{0.0097} & \textbf{0.0000} & \textbf{0.0096} & \textbf{0.0000} & \textbf{0.0000} & \textbf{0.0000} \\
     & Median & 0.1823 & 0.9991 & 0.1786 & n/a & n/a & n/a \\
     \hline
    \multirow{5}{*}{\texttt{FAv2}} & Min & 0.6483 & 0.9993 & 0.6479 & n/a & n/a & n/a \\
     & Max & 0.7664 & 0.9994 & 0.7662 & n/a & n/a & n/a \\
     & Mean & \textbf{0.7077} & \textbf{0.9994} & \textbf{0.7075} & \textbf{0.0000} & \textbf{0.0000} & \textbf{0.0000} \\
     & StDev & \textbf{0.0230} & $\mathbf{<0.0001}$ & \textbf{0.0230} & \textbf{0.0000} & \textbf{0.0000} & \textbf{0.0000} \\
     & Median & 0.7093 & 0.9994 & 0.7091 & n/a & n/a & n/a \\
     \hline
    \multirow{5}{*}{\texttt{BA}} & Min & 0.6457 & 0.9993 & 0.6457 & n/a & n/a & n/a \\
     & Max & 0.9889 & 0.9996 & 0.9889 & n/a & n/a & n/a \\
     & Mean & \textbf{0.9566} & \textbf{0.9995} & \textbf{0.9570} & \textbf{0.0000} & \textbf{0.0000} & \textbf{0.0000} \\
     & StDev & \textbf{0.0345} & \textbf{0.0001} & \textbf{0.0346} & \textbf{0.0000} & \textbf{0.0000} & \textbf{0.0000} \\
     & Median & 0.9647 & 0.9995 & 0.9651 & n/a & n/a & n/a \\
    \hline
    \end{tabularx}
\end{table}     

As can be seen in the table, the strongest equivalence regarding the statistical similarity metrics was identified between the \texttt{PSO} and \texttt{accPSO}, where the ''very strong'' correlation was detected by all the observed metrics. Moreover, the correlation was even stronger than by identifying the equivalence between the same algorithms, but an opposite control/controlled algorithm (i.e., \texttt{accPSO/PSO}). Interestingly, the ''very strong'' correlation was also identified between the \texttt{PSO} and \texttt{BA}, while the ''very weak'' correlation was detected regarding the cosine similarity and Spearman correlation coefficient between the \texttt{PSO} and original \texttt{FA}. The ''strong'' correlation regarding both the mentioned statistical similarity metrics was also identified by the \texttt{FAv2}.

On the other hand, the results according to the ML similarity metrics did not indicate any correlation between the control \texttt{PSO} and controlled NI algorithms from the set $\mathit{NIA}^{(\text{CS-2})}\backslash\texttt{accPSO}$.

\subsubsection{CS-3: Visualization of the results}
\begin{figure}[htb]
    \centering
    \begin{subfigure}{0.48\linewidth}
    \includegraphics[width=1.0\linewidth]{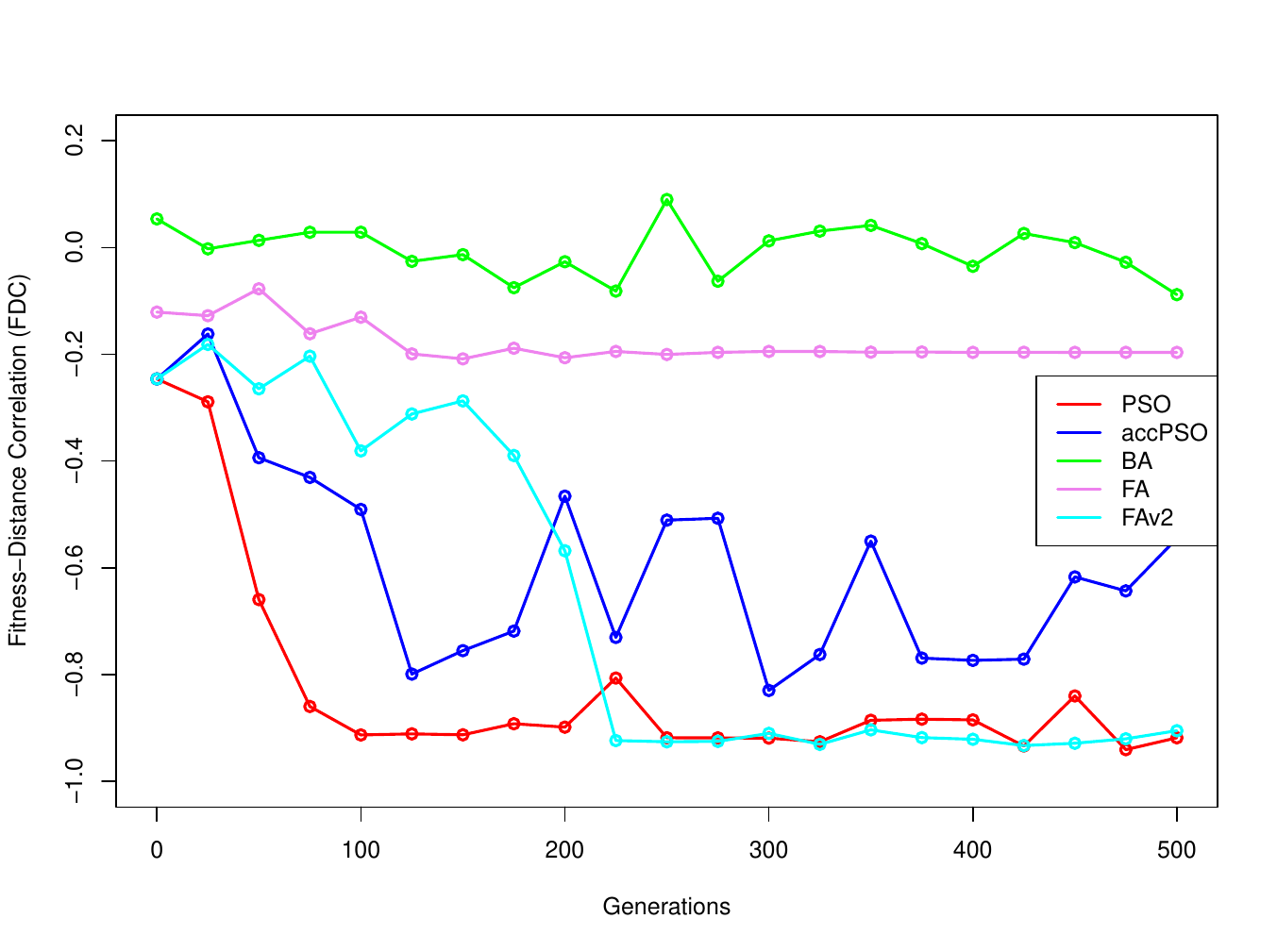} 
    \caption{Fitness/Distance Correlation (FDC).}
    \label{fig:cg-11}
    \end{subfigure}
    \begin{subfigure}{0.48\linewidth}
    \includegraphics[width=1.0\linewidth]{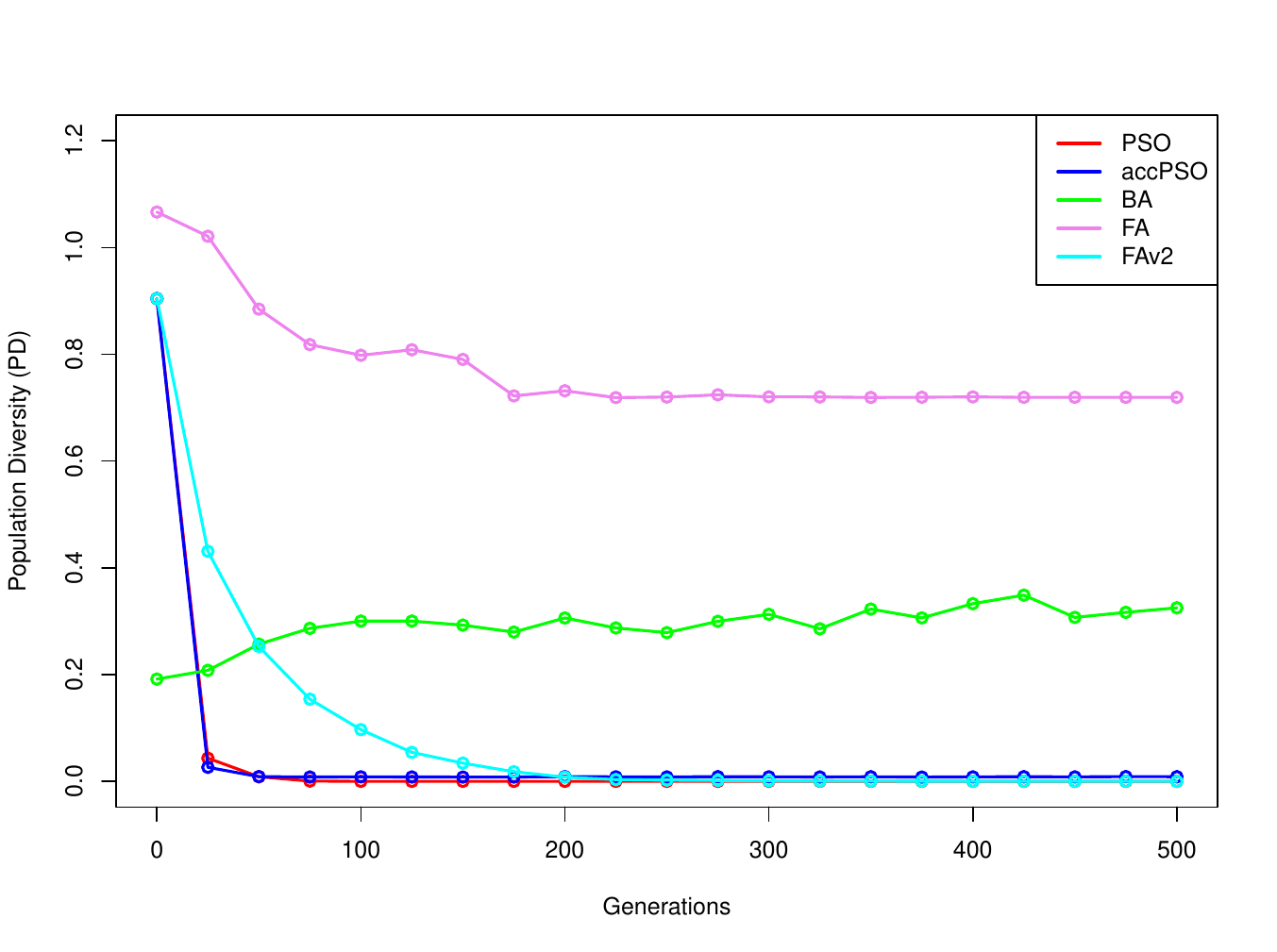}
    \caption{Population Diversity (PD).}
    \label{fig:cg-12}
    \end{subfigure}
    \begin{subfigure}{0.48\linewidth}
    \includegraphics[width=1.0\linewidth]{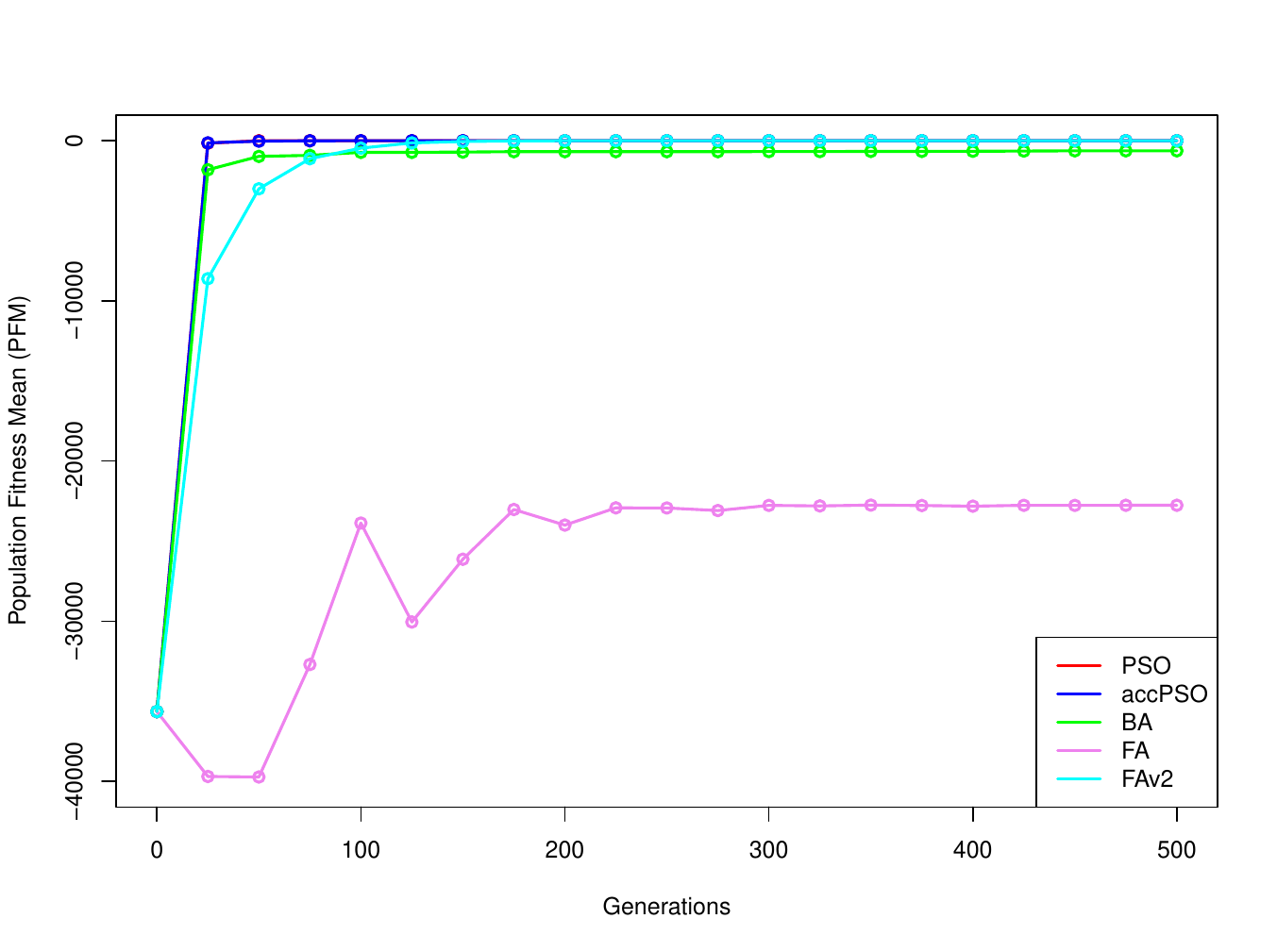} 
    \caption{Population fitness mean (PFM).}
    \label{fig:cg-13}
    \end{subfigure}
    \begin{subfigure}{0.48\linewidth}
    \includegraphics[width=1.0\linewidth]{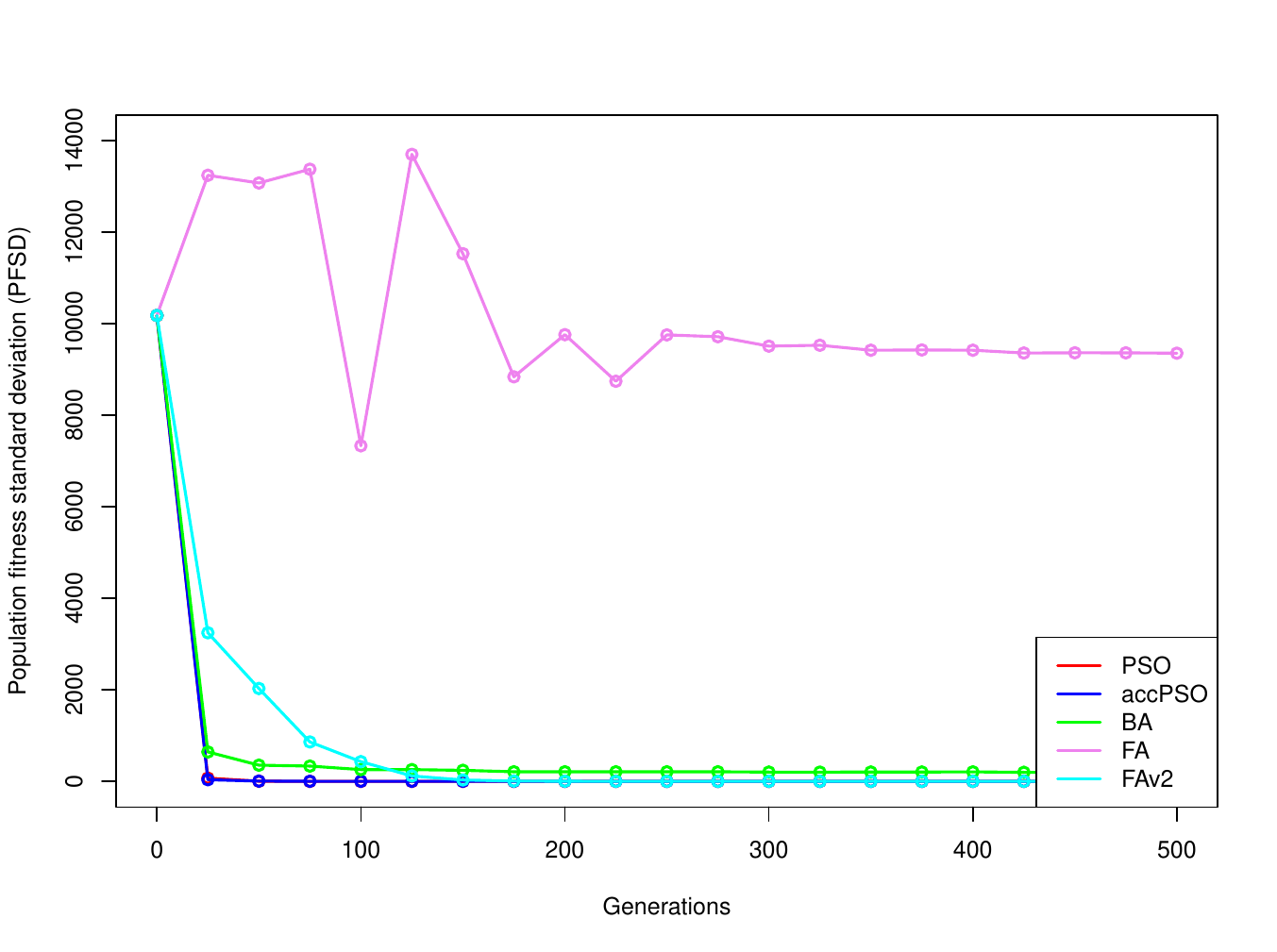}
    \caption{Population fitness standard deviation PFSD).}
    \label{fig:cg-14}
    \end{subfigure}
    \caption{Population descriptive metrics.}
    \label{fig:cg-1}
\end{figure}

\begin{figure}[htb]
    \centering
    \begin{subfigure}{0.48\linewidth}
    \includegraphics[width=1.0\linewidth]{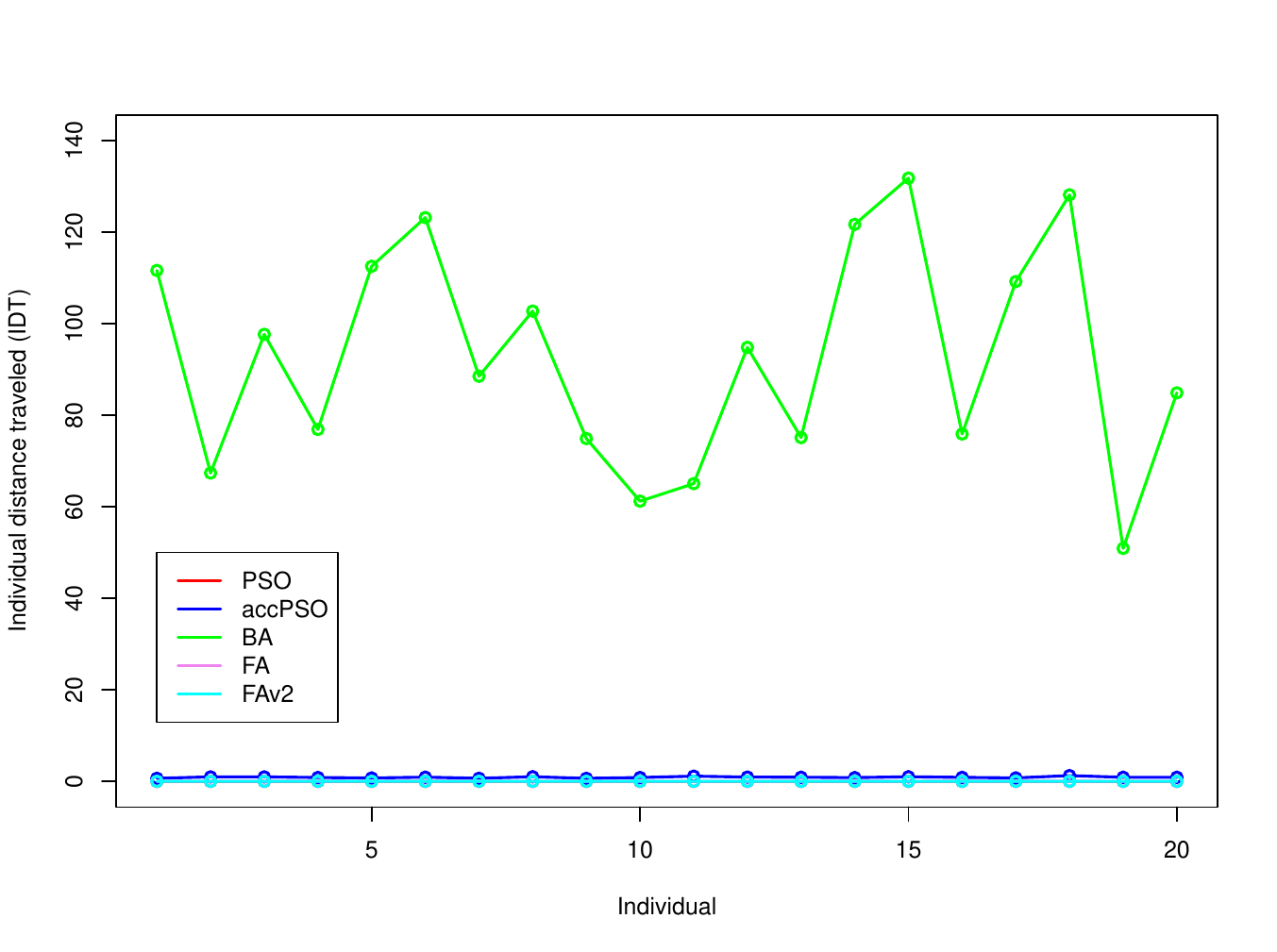} 
    \caption{Individual distance traveled (IDT).}
    \label{fig:cg-21}
    \end{subfigure}
    \begin{subfigure}{0.48\linewidth}
    \includegraphics[width=1.0\linewidth]{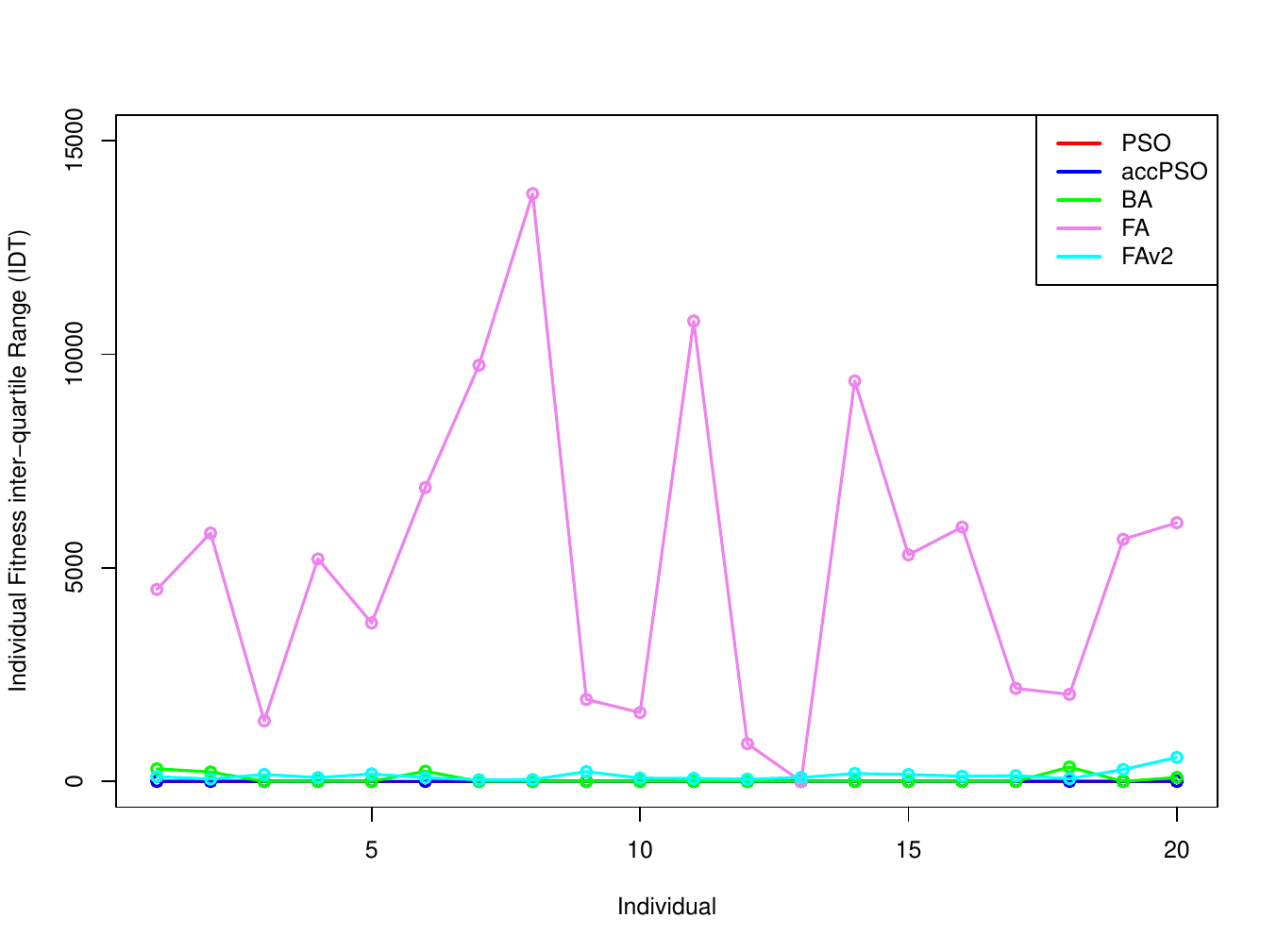}
    \caption{Individual fitness inter-quartile range.}
    \label{fig:cg-22}
    \end{subfigure}
    \begin{subfigure}{0.48\linewidth}
    \includegraphics[width=1.0\linewidth]{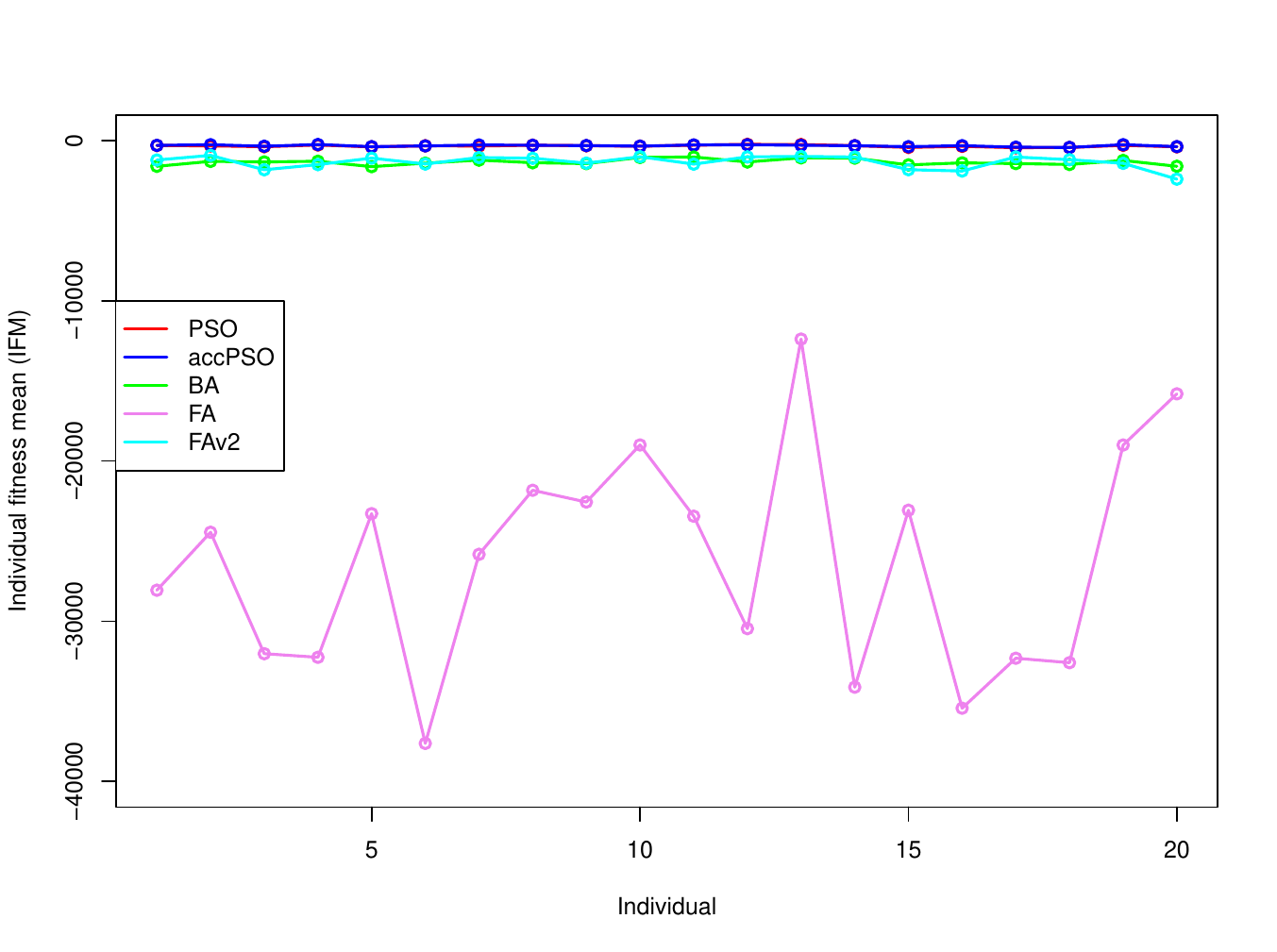} 
    \caption{Individual fitness mean (IFM).}
    \label{fig:cg-23}
    \end{subfigure}
    \begin{subfigure}{0.48\linewidth}
    \includegraphics[width=1.0\linewidth]{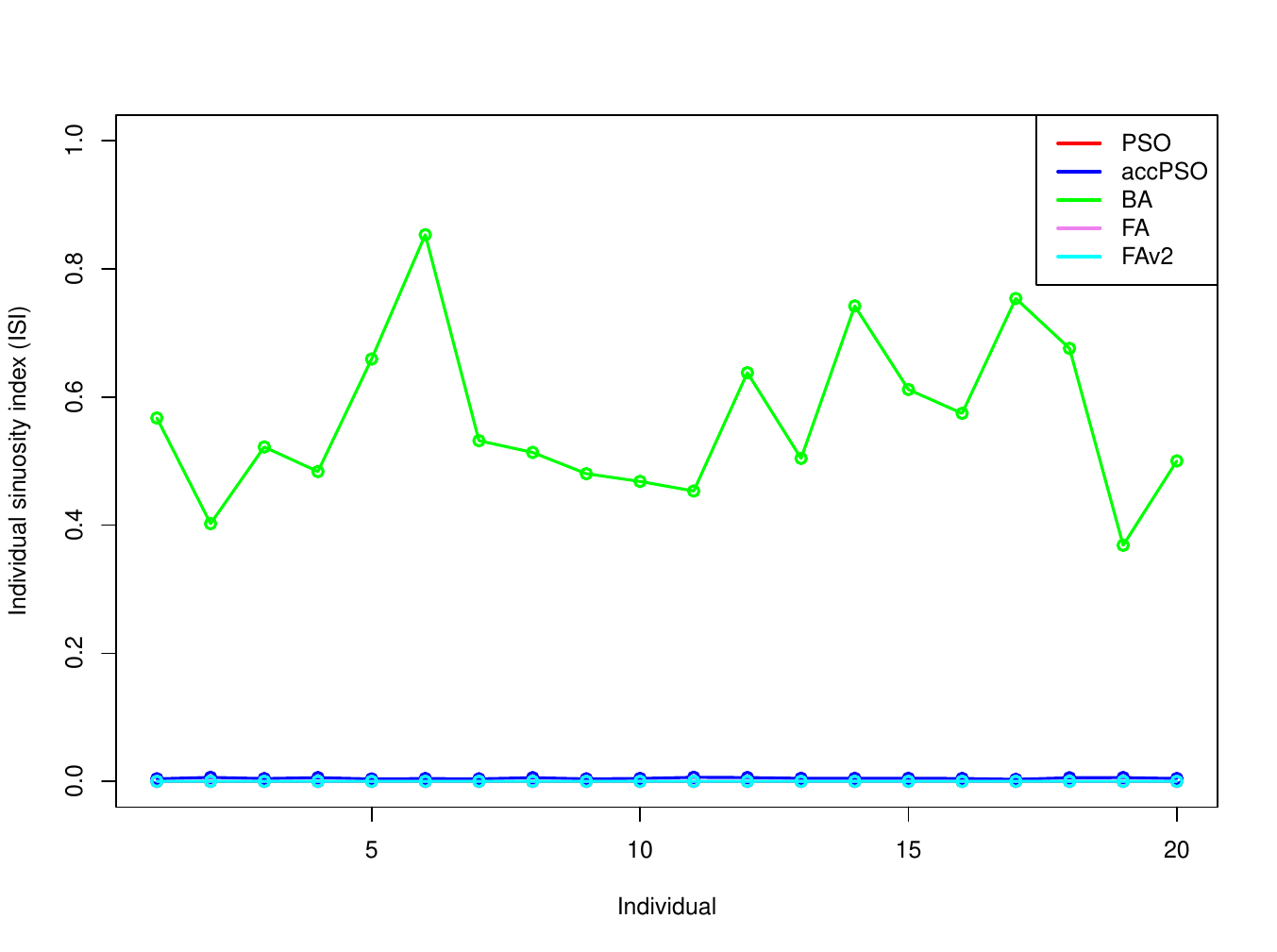}
    \caption{Individual sinuosity index (ISI).}
    \label{fig:cg-24}
    \end{subfigure}
    \caption{Individual descriptive metrics.}
    \label{fig:cg-2}
\end{figure}

In order to demonstrate the behavior of the definite quality measures during the typical evolutionary run by various NI algorithms, the population metrics are represented graphically in Fig.~\ref{fig:cg-1}, while the individual metrics in Fig.~\ref{fig:cg-2}. The data for plotting the charts were aggregated during the Case study~3 by executing the NI metaheuristics initialized with the seed of random generator set to one. 

Fig.~\ref{fig:cg-11} depicts the behavior of an FDC metric by the observed NI algorithms. 
The FDC is a measure for predicting the performance of EAs by assessing the relation between fitness and distance from the optimal solution. In line with this, the value of the value $\approx -1$ means that the problem is easy to be solved for the observed NI algorithm. On the other hand, the problem is hard for the NI algorithm when this value is close to zero. As can be seen from the figure, the problem is easy for the \texttt{PSO} and \texttt{FAv2}, while it is hard for the \texttt{BA} and \texttt{FA} algorithms.

A PD is the variety of genetic material in the current population maintained by the particular NI algorithm. Too low population diversity shows the higher selection pressure that can lead to premature convergence. The steady decline of population diversity over the generations is a sign of a normal exploitation phase. This behavior is observable in Fig.~\ref{fig:cg-12} by the algorithms \texttt{PSO}, \texttt{accPSO}, and \texttt{FAv2}, while the higher population diversity was maintained by the algorithms \texttt{BA} and \texttt{FA}.

A lot of information around the state of the evolutionary search process can be used in the monitoring, control, and analysis of the PFM and PFSD. For instance, increasing mean monotonically is a sign of finding the better solutions steadily by the definite NI algorithm. On one hand, the high mean with the lower standard deviation shows that the population is homogeneous, while the evolutionary search process exploits mainly the search space. On the other hand, the moderate mean and high standard deviation indicates ongoing exploration supported by a mixture of good and mediocre individuals. In Figs.~\ref{fig:cg-13}-\ref{fig:cg-14}, increasing of the means and decreasing of the standard deviations are indicated by all the observed NI algorithms except the \texttt{FA}.

The IDT metric is suitable to track how far the genotype is moved from one generation to the next by each individual on average. A high average distance indicates that the NI algorithm's search process is in the exploration phase, while the lower average distances show that the population is either converged, or is stuck into the local optimum. The mixed distances, where some individuals move more in  the search space and the other only barely, signal balancing between exploration and exploitation by the search process. In Fig.~\ref{fig:cg-21} 
it is evident that the mixed distances are indicated only in the \texttt{BA} population, while the individuals in the other algorithm's populations are moved only slightly on average.

The IQIQR is a compact, robust indicator of how much the middle half of your population differs in quality. In line with this, a large IQIQR is connected with the higher population diversity, which is also a sign that the evolutionary search process is in the exploratory phase. On the other hand, a smaller IQIQR shows that the population converges, possibly at risk of getting stuck into the local optimum. Fig.~\ref{fig:cg-22} depicts that the higher values of the IQIQR metric was obtained by the original \texttt{FA} (i.e., the exploration phase), while the values observed by the other algorithms are slightly lower, but homogeneous among all the population members (i.e., convergence).

The IFM of all the individuals in a generation is one of the most straightforward summary metrics used to monitor the behavior of the NI algorithms. It condenses the overall quality of the current population into a single number that is interpreted typically as follows: Increasing mean means that the population members are improving on average. On the other hand, a plateaued or decreasing mean indicates stagnation, possible premature convergence, or ineffective operators. As can be seen in Fig.~\ref{fig:cg-23}, the the IFM values demonstrated the stagnation or possible premature convergence by the original \texttt{FA} only, while improving values were detected by all the other observed NI algorithms.

The ISI is a scalar measure that captures how indirect a solution’s trajectory is within the search space of an NI algorithm in respect to its initial position.  The measure offers a concise, descriptive lens on how each population member traverses the search space, helping to diagnose the algorithm's behavior, tune parameters, and understand the interaction between the operators and problem topology. As can be observed in Fig.~\ref{fig:cg-24}, these traverses are more homogeneous, focused in specific regions of the search space by all the observed NI algorithms, except the \texttt{BA}, where these trajectories indicate that the search space was explored more randomly by this algorithm. 

\subsubsection{CS-3: Accepting/rejecting the hypothesis}
In general, the framework revealed that creating the conditions for achieving a strong equivalence between NI metaheuristics algorithms is hard. Moreover, the strong equivalence between the PSO and FA can be increased by increasing the cosine similarity metric if an improved FAv2 is applied. In summary, it seems that we fall into the trap of comparing apples and oranges: On the one hand, there are variation operators that, theoretically, are not distinguished from algorithm to algorithm too much, while, practically, each algorithm implements different algorithm's components on the other. Additionally, when we also consider the stochastic nature of the algorithms in the study, as well as the limited computational resources, this makes the conditions for fulfilling the equivalence theorem become even harder.

Consequently, \textbf{Hypothesis~\ref{hypo:3}} can only be accepted conditionally due to the limitations of the computer resources used.

\subsection{Case Study 4: The BA is PSO}

The \textbf{Case study~4} is devoted to show how can the framework helped us by accepting/rejecting \textbf{Hypothesis~\ref{hypo:4}} that states universally: ''The algorithm A is B'', where as algorithm~A was taken the \texttt{BA}, and as algorithm~B the \texttt{PSO}. In the remainder of the paper, the steps required for hypothesis testing are presented in detail.

\subsubsection{CS-4: Selection of the algorithm's test set}
The set of NI metaheuristic algorithms in the case study was defined similarly as in the last case study, i.e., $\mathit{NIA}^{(\text{CS-4})}=\{\texttt{accPSO,PSO,FA,FAv2,BA}\}$. The original \texttt{BA} algorithm was taken as a control algorithm using the following parameter settings: $A=1.0$, and $\gamma=0.1$ by generating the reference feature vectors. 

\subsubsection{CS-4: Results of the parameter setting analysis}
The results of the best parameter setting as found by the \texttt{Meta-DE} together with the corresponding fitness function values are illustrated in Table~\ref{tab:cs4}.
\begin{small}
\begin{table}[htb]
    \caption{The results of the \texttt{Meta-DE} by parameter setting analysis in CS-4.}
    \label{tab:cs4}
    \centering
    \begin{tabularx}{\textwidth}{l|l|l|rrrrr}
    \hline
    Role & Algorit. & Metric & Min & Max & Mean & StDev & Median\\
    \midrule
    \multirow{3}{*}{\rotatebox{90}{Control}} & \multirow{3}{*}{\texttt{BA}} & $A$ & 1.0000 & 1.0000 & 1.0000 & 0.0000 & 1.0000 \\
     & & $\gamma$ & 0.1000 & 0.1805 & 0.1148 & 0.0199 & 0.1041 \\
     \cmidrule{3-8}
     & & Fitness & -1.5632 & -0.3465 & \textbf{-1.0069} & \textbf{0.2497} & -1.0332 \\
     \cmidrule{1-8}
    \multirow{18}{*}{\rotatebox{90}{Controlled}} & \multirow{3}{*}{\texttt{accPSO}} & $\alpha$ & 1.0000 & 1.0000 & 1.0000 & 0.0000 & 1.0000 \\
     & & $\beta$ & 0.3911 & 0.9629 & 0.7815 & 0.0569 & 0.7869 \\
     \cmidrule{3-8}
     & & Fitness & -0.2243 & -0.0557 & \textbf{-0.1375} & \textbf{0.0341} & -0.1412 \\
     \cmidrule{2-8}
     & \multirow{4}{*}{\texttt{PSO}} & $w$ & \textbf{0.4000} & \textbf{0.9000} & \textbf{0.7095} & 0.1539 & 0.7639 \\
     & & $c_1$ & 1.503 & 2.500 & 2.4452 & 0.1585 & 2.5000 \\
     & & $c_2$ & \textbf{1.5000} & \textbf{2.5000} & \textbf{1.7959} & 0.3173 & 1.6994 \\
     \cmidrule{3-8}
     & & Fitness & -70.5013 & $<-0.0001$ & \textbf{-8.1498} & \textbf{13.3765} & -2.3702 \\
     \cmidrule{2-8}
     & \multirow{4}{*}{\texttt{FA}} & $\alpha$ & 0.6919 & 1.0000 & 0.9540 & 0.0445 & 0.9662 \\
     & & $\beta$ & \textbf{0.5000} & \textbf{1.5000} & \textbf{1.1339} & 0.3383 & 1.2003 \\
     & & $\gamma$ & \textbf{0.1000} & \textbf{1.0000} & \textbf{0.3973} & 0.3146 & 0.2990 \\
     \cmidrule{3-8}
     & & Fitness & -9792.5300 & -3621.3300 & \textbf{-6505.5861} & \textbf{1291.5688} & -6387.4500 \\
     \cmidrule{2-8}
     & \multirow{4}{*}{\texttt{FAv2}} & $\alpha$ & 0.1129 & 0.6902 & 0.3587 & 0.1463 & 0.3567 \\
     & & $\beta$ & \textbf{0.5000} & \textbf{1.5000} & \textbf{0.9746} & 0.3884 & 0.8890 \\
     & & $\gamma$ & \textbf{0.1000} & \textbf{1.0000} & \textbf{0.5957} & 0.3266 & 0.6109 \\
     \cmidrule{3-8}
     & & Fitness & -0.2699 & -0.0040 & \textbf{-0.0823} & \textbf{0.0608} & -0.0726 \\
    \hline     
    \end{tabularx}
\end{table}
\end{small}

As is obvious in the table, the \texttt{Meta-DE} control system revealed the high variances by reconstructing the parameter $\gamma$ of the \texttt{BA}. Interestingly, the parameter $\alpha$ in the controlled \texttt{accPSO} was matched exactly by the control \texttt{BA}, while the parameters $w$ and $c_2$ in the \texttt{PSO}, and the parameters $\beta$ and $\gamma$ in both \texttt{FA} versions were searched within the whole domain of feasible values by the \texttt{Meta-DE}. All the parameter values are embolden in the table.

Regarding the fitness values obtained by the search for the strongest equivalence between the control and set of controlled NI algorithms, the narrowed results to the control \texttt{BA} were reported by the \texttt{accPSO} and the \texttt{PSO} algorithms, while the worst were obtained by the original \texttt{FA}, on average.

\subsubsection{CS-4: Results of identifying the strong equivalence of the NI algorithms}
The results of the statistical analysis for identifying the strong equivalence of the control algorithm \texttt{BA} and controlled algorithms drawn from the set $\mathit{NIA}^{(\text{CS-4})}\backslash\texttt{BA}$ are depicted in Table~\ref{tab:cs4-detail}. 

\begin{table}[htb]
    \caption{The results of the statistical analysis according to the similarity metrics for identifying the strong equivalence between the \texttt{BA} as the control algorithm in CS-4.}
    \label{tab:cs4-detail}
    \centering
    \begin{tabularx}{\textwidth}{l|l|rrrrrr}
    \hline
    Algorithm & Parameter & $Sim_{\cos}$ & $Sim_{\text{SMAPE}}$ & $\rho$ & $Sim_{\text{kNN}}$ & $Sim_{\text{SVM}}$ & $Sim_{\text{RF}}$ \\
    \midrule
    \multirow{5}{*}{\texttt{accPSO}} & Min & 0.8605 & 0.9996 & 0.8607 & n/a & n/a & n/a \\
     & Max & 0.9575 & 0.9998 & 0.9573 & n/a & n/a & n/a \\
     & Mean & \textbf{0.9069} & \textbf{0.9997} & \textbf{0.9050} & \textbf{0.0000} & \textbf{0.0000} & \textbf{0.0000} \\
     & StDev & \textbf{0.0142} & \textbf{0.0000} & \textbf{0.0147} & \textbf{0.0000} & \textbf{0.0000} & \textbf{0.0000} \\
     & Median & 0.9064 & 0.9997 & 0.9042 & n/a & n/a & n/a \\
     \hline
    \multirow{5}{*}{\texttt{PSO}} & Min & 0.6173 & 0.9993 & 0.6108 & n/a & n/a & n/a \\
     & Max & 0.8872 & 0.9997 & 0.8879 & n/a & n/a & n/a \\
     & Mean & \textbf{0.7270} & \textbf{0.9994} & \textbf{0.7264} & \textbf{0.0000} & \textbf{0.0000} & \textbf{0.0000} \\
     & StDev & \textbf{0.0614} & $\mathbf{<0.0001}$ & \textbf{0.0623} & \textbf{0.0000} & \textbf{0.0000} & \textbf{0.0000} \\
     & Median & 0.7187 & 0.9994 & 0.7213 & n/a & n/a & n/a \\
    \hline
    \multirow{5}{*}{\texttt{FA}} & Min & 0.2319 & 0.9991 & 0.2254 & n/a & n/a & n/a \\
     & Max & 0.5247 & 0.9991 & 0.5030 & n/a & n/a & n/a \\
     & Mean & \textbf{0.4364} & \textbf{0.9991} & \textbf{0.4171} & \textbf{0.0000} & \textbf{0.0000} & \textbf{0.0000} \\
     & StDev & \textbf{0.0507} & $\mathbf{<0.0001}$ & \textbf{0.0486} & \textbf{0.0000} & \textbf{0.0000} & \textbf{0.0000} \\
     & Median & 0.4485 & 0.9991 & 0.4279 & n/a & n/a & n/a \\
     \hline
    \multirow{5}{*}{\texttt{FAv2}} & Min & 0.7074 & 0.9993 & 0.7063 & n/a & n/a & n/a \\
     & Max & 0.9798 & 0.9998 & 0.9797 & n/a & n/a & n/a \\
     & Mean & \textbf{0.9413} & \textbf{0.9997} & \textbf{0.9409} & \textbf{0.0000} & \textbf{0.0000} & \textbf{0.0000} \\
     & StDev & \textbf{0.0375} & \textbf{0.0001} & \textbf{0.0379} & \textbf{0.0000} & \textbf{0.0000} & \textbf{0.0000} \\
     & Median & 0.9508 & 0.9997 & 0.9506 & n/a & n/a & n/a \\
     \hline
    \end{tabularx}
\end{table}

As is evident from the table, the strongest equivalence was identified by the \texttt{FAv2} and then the \texttt{accPSO} algorithms. In both cases, the ''very strong'' correlation was detected by the statistical similarity measures $Sim_{\cos}$ and the Spearman correlation coefficient $\rho$. The ''strong'' correlation was identified between the \texttt{BA} and the \texttt{PSO} algorithms, and the ''moderate'' between the \texttt{BA} and \texttt{FA}. 

The strong identity regarding the ML similarity measures did not indicate any correlation between the control \texttt{BA} and any other controlled NI algorithms from the set $\mathit{NIA}^{(\text{CS-4})}\backslash\texttt{BA}$ .

\subsubsection{CS-4: Accepting/rejecting the hypothesis}
Although the framework discovered that \textbf{Hypothesis-\ref{hypo:4}} cannot be accepted in general, the speculation that both the BA and PSO algorithms are strongly equivalent, however, could be approved only partially: If, for example, we exclude the \texttt{GRW} component (by setting the pulse rate to $\gamma = 0.0$) and the \texttt{SA} component (by setting the loudness to $A_0=1.0$), and widen the search space size of the PSO by changing the feasible domain of values for the social inertia coefficient from $c1\in[1.5,2.5]$ to $c1\in[0.0,2.5$], then the algorithms can behave equivalently.

\subsection{Discussion}\label{sec12}
The purpose of the study was to develop a framework for identifying a strong equivalence between NI metaheuristics. In line with this, the strong equivalence is defined according to which two NI algorithms are equivalent, if they produce the results in the phenotype space, and exhibit the behavior by the exploring the search space that does not differ for less than 1~\% regarding the cosine similarity metric. This goes against most of the existing studies that prove this similarity, not generally, but on a theoretical basis, without any strong evidence based on the achieved results. 

It holds commonly that the majority of the SI-based algorithms are found on loosely-coupled foundations that can be interpreted in the sense of inspiration from the nature seemingly. In our opinion, the main reason for the disorder can be found in the fact that there does not exist a clear classification of these algorithms. While the classification of EAs was made by \citet{baeck1993overview} already in 1993, the commonly accepted classification does not exist inside the SI-based community. 

Let us remind, that the classification of EA algorithms bases on the different representation of individuals, as follows: (1) Genetic Algorithms (GA) working on binary vectors~\cite{goldberg1989genetic}, (2) Genetic Programming (GP) ~\cite{koza1992genetic} working on programs written in the Lisp programming language~\cite{koza1992genetic}, (3) Evolution Strategies~\cite{eigen1973ingo,beyer2002evolution} and Differential Evolution working on real-valued vectors~\cite{storn1997differential}, while (4) Evolutionary Programming (EP) working on finite-state machines~\cite{fogel1966artificial}. 

On the other hand, the SI-based algorithm community has rarely posted a rigorous classification. Indeed, most algorithms of this family operate using the representation of individuals as the real-valued vectors, and, consequently, these algorithms cannot be classified based on the representation of individuals. They are easier to classify regarding the employed metaphor. In our opinion, the potential solution of this problem could be to define the general framework of the SI-based metaheuristic algorithms without any NI metaphors as already proposed, for instance, by \citet{Fister2016towards} ten years ago. In accordance with this, the new metaheuristic algorithms with clear design would be implemented, that differ between each other by applying different adaptation and hybridization methods similar as in the EA community. 

In searching for how original are the ''novel''  metaheuristic algorithms, the proposed framework can be very helpful. In summary, the results have revealed that it can: (1) recognize two similar NI metaheuristic algorithms according to the strong equivalence theorem, (2) be helpful in developing and analyzing the new NI metaheuristic components, (3) estimate the quality of the NI metaheuristics in the phenotype space, and exhibit those algorithm components that affect the obtained results the most crucially, and (4) answer to the hypothesis ''The A is B'' in the sense of the equivalence theorem, in general. However, all these conclusions hold inside the time and space limitations of the used computational resources.     

\section{Conclusion}\label{sec:5}
This paper proposes a framework for identifying the strong equivalence between NI metaheuristics. In line with this, descriptive metrics are defined for measuring the behavior of these algorithms at the individual as well as the population level. The results of collecting these metrics during the whole evolutionary runs are represented in feature vectors, that enter into the comparison regarding the statistical similarity metrics and ML classifiers, in order to estimate how equivalent two NI metaheuristics really are. In each comparison step, the control algorithm is defined for generating the reference feature vectors, while the set of controlled NI metaheuristic algorithms are defined capable of generating the feature vectors, with which the reference ones are compared using the statistical similarity metrics and ML classifiers. To automate the process of searching for the most similar feature vectors the \texttt{Meta-DE} is used, that is capable of tuning the algorithm's parameters of the controlled metaheuristics.

Using the proposed framework four case studies were conducted on a set of seven metaheuristics, in order to show: (1) the so-called accelerated \texttt{accPSO}/\texttt{accFA} are strong equivalent algorithms, although they are founded on different metaphors, (2) the NI metaphor provides the inspiration for implementation of only a few metaheuristic components, (3) creating conditions in which we could show that two NI metaheuristics behave strongly equivalently, is hard to identify, due to using unique algorithm's components and limited computer resources, and (4) to assert that ''The algorithm A is B'' is uncritical, as can be seen by using the strong equivalent theorem.

The potential directions for the future work in this domain are huge. Let us mention only some: (1) due to the complexity of the framework, some steps should be more automated, in order to draw its application nearer to the regular user, (2) the framework should be connected with methods of XAI, (3) due to its time and space complexity, the framework should be moved to the super computer environment, and (4) to broaden its applicability, this should be employed on a bigger set of NI metaheuristic algorithms.

\end{document}